\pdfoutput=1

\documentclass[3p]{elsarticle}

\usepackage{amssymb}
\usepackage{amsmath}
\usepackage{graphicx}
\usepackage{color}
\usepackage{caption}
\usepackage{subcaption}
\usepackage{booktabs}

\DeclareMathOperator*{\argmax}{arg\,max}

\journal{Computer Vision and Image Understanding}

\begin{document}

\begin{frontmatter}



\title{No Spare Parts: Sharing Part Detectors for Image Categorization}

\author[label1]{Pascal Mettes\corref{cor1}}
\ead{P.S.M.Mettes@uva.nl}
\author[label1,label2]{Jan C. van Gemert}
\author[label1,label3]{Cees G. M. Snoek}
\address[label1]{University of Amsterdam, Science Park 904, Amsterdam, the Netherlands}
\address[label2]{Delft University of Technology, Mekelweg 4, Delft, the Netherlands}
\address[label3]{Qualcomm Research Netherlands, Science Park 400, Amsterdam, the Netherlands}
\cortext[cor1]{Corresponding author}

\begin{abstract}
This work aims for image categorization by learning a representation of discriminative parts.
{\color{black}Different from most existing part-based methods, we argue that parts are naturally shared between image categories and should be modeled as such.}
We motivate our approach with a quantitative and qualitative analysis by backtracking where selected parts come from. Our analysis shows that in addition to the category parts defining the category, the parts coming from the background context and parts from other image categories improve categorization performance.
Part selection should not be done separately for each category, but instead be shared and optimized over all categories.
To incorporate part sharing between categories, we present an algorithm based on AdaBoost to optimize part sharing and selection, as well as fusion with the global image representation.
{\color{black}With a single algorithm and without the need for task-specific optimization,} we achieve results competitive to the state-of-the-art on object, scene, and action categories, further improving over deep convolutional neural networks and alternative part representations. 
\end{abstract}

\begin{keyword}
Image categorization \sep Discriminative parts \sep Part sharing



\end{keyword}

\end{frontmatter}


\section{Introduction}
\label{sec:intro}

In this work, we aim to categorize images into their object, scene, and action category. Image categorization has been studied for decades and tremendous progress has been made~\cite{lazebnik2006beyond,sanchez13,sivic2003video,sande10,gemert2010visual,zhang2007local}, 
most recently by the introduction of very deep convolutional neural networks~\cite{krizhevsky12,simonyan15,szegedy14,zeiler14}. 
These networks learn to categorize images from examples by back-propagating errors through stacked layers of convolutional filters pooled over image regions. Despite their implicit local nature, deep nets result in a global scene representation only. Thereby, ignoring known benefits of explicitly encoding local image blocks, i.e. discriminative power~\cite{doersch13,zuo14}, image interpretation~\cite{doersch12,freytag14}, and complementarity~\cite{juneja13,zuo14}. In this paper we make a case for sharing parts for image categorization, studying \emph{what} parts to consider, \emph{which} parts to select and \emph{how} to share them between categories.

{\color{black}The notion of sharing has been well studied in data mining~\cite{rematas2015dataset,yuan2008mining}. These works repeatedly show that sharing boosts classification performance and provides connections between co-occurring elements. Motivated by these examples, we investigate sharing of parts when learning part-representations for image categorization.} 

\begin{figure*}[t]
\centering
\begin{subfigure}[b]{0.475\linewidth}
\centering
\begin{minipage}{0.49\textwidth}
\begin{center} {Image} \end{center}
\vspace{-0.35cm}
\includegraphics[width=\linewidth]{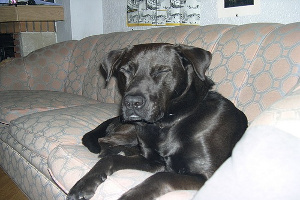}
\end{minipage}
\begin{minipage}{0.49\textwidth}
\begin{center} {Own responses} \end{center}
\vspace{-0.35cm}
\includegraphics[width=\linewidth]{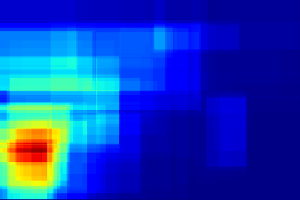}
\end{minipage}
\vspace{0.35cm}\\
\begin{minipage}{0.49\textwidth}
\begin{center} {Other responses} \end{center}
\vspace{-0.35cm}
\includegraphics[width=\linewidth]{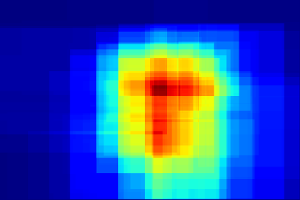}
\end{minipage}
\begin{minipage}{0.49\textwidth}
\begin{center} {Context responses} \end{center}
\vspace{-0.35cm}
\includegraphics[width=\linewidth]{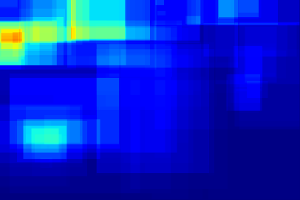}
\end{minipage}
\caption{\texttt{Sofa} category.}
\label{fig:share-sofa}
\end{subfigure}
\hspace{0.3cm}
\begin{subfigure}[b]{0.475\linewidth}
\centering
\begin{minipage}{0.49\textwidth}
\begin{center} {Image} \end{center}
\vspace{-0.35cm}
\includegraphics[width=\linewidth]{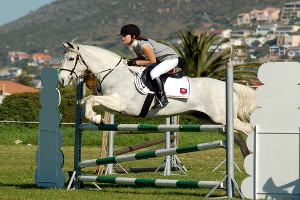}
\end{minipage}
\begin{minipage}{0.49\textwidth}
\begin{center} {Own responses} \end{center}
\vspace{-0.35cm}
\includegraphics[width=\linewidth]{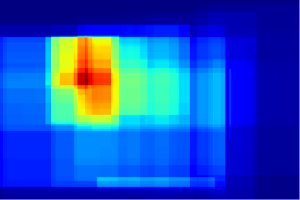}
\end{minipage}
\vspace{0.35cm}\\
\begin{minipage}{0.49\textwidth}
\begin{center} {Other responses} \end{center}
\vspace{-0.35cm}
\includegraphics[width=\linewidth]{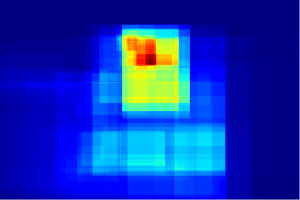}
\end{minipage}
\begin{minipage}{0.49\textwidth}
\begin{center} {Context responses} \end{center}
\vspace{-0.35cm}
\includegraphics[width=\linewidth]{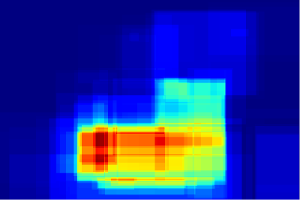}
\end{minipage}
\caption{\texttt{Horse} category.}
\label{fig:share-horse}
\end{subfigure}
\caption{Shared part responses of our method for an example of \texttt{sofa} (a) and \texttt{horse} (b). We show that image categories do not only benefit from own category part responses, but also from responses of other categories (e.g. \texttt{dog} for \texttt{sofa} and \texttt{person} for \texttt{horse}) and from context (e.g. horizontal bars for \texttt{horse}).}
\label{fig:share-example}
\end{figure*}

Consider Figure~\ref{fig:share-example}, where examples of image categories \texttt{sofa} and \texttt{horse} utilize parts from their own category as well as parts from other categories and the background context. When a classifier is trained exclusively using parts from its own category, relevant information is missed~{\color{black}\cite{azizpour2015spotlight,parizi15}}. As illustrated in Figure~\ref{fig:share-example}, for object categories such as \texttt{sofa}, it is informative to use parts from \texttt{cat} and \texttt{dog} categories as well as parts from \texttt{sofa} itself~{\color{black}\cite{azizpour2015spotlight,parizi15}}. By giving the \texttt{sofa} classifier access to \texttt{dog} and \texttt{cat} training images, the recognition of \texttt{sofa} is improved even though these images may not contain a \texttt{sofa} at all. Even when global image categories differ, their representation can share similar parts and should thus be modeled as such~{\color{black}\cite{juneja13,parizi15}}.

To obtain insight in part sharing, we track where a part comes from. We define three types of part-origin. \emph{Own}: parts coming from the defining image category; \emph{Other}: parts coming from other labeled categories; \emph{Context}: parts not overlapping with labeled categories. Our analysis shows that the global image representation captures the scene; that Own parts benefit out-of-context and small images (e.g. \texttt{sofa} in the grass); Other parts aid for co-occurring categories (e.g. \texttt{person} on a \texttt{horse}) and Context parts help to recognize supporting structures (e.g. fence for \texttt{sheep} or \texttt{cows}).

Several recent works have focused on local parts as representation for image classification~\cite{bossard14,doersch13,doersch12,juneja13} by performing part selection for each category separately. These works follow a common four-step pipeline. First, a set of potentially interesting parts is proposed. Second, a selection algorithm discovers the most discriminative parts. Third, classification is performed on the part-based representations. And fourth, fusion with a global image representation. These four steps have shown to yield robust and complementary image representations~\cite{bossard14,doersch13,zuo2015exemplar}. However, by separating the part selection, category classification, and fusion steps into disjoint steps these works leave room for a number of improvements. 
Most notably, (1) the models do not share parts and perform part selection for each category independently, {\color{black}as also observed by Parizi et al.~\cite{parizi15}}, 
(2) the models use different objective functions for the part selection step and the final classification step,
and (3) the models perform part selection independently of the global representation with which they are fused.
As we will show, these drawbacks result in sub-optimal image categorization results. 

We make three contributions in this paper:
(i) {\color{black} we establish that three part types are relevant for image categorization, which are all naturally shared between categories when learning a part-representation for image categorization};
(ii) {\color{black} we present an algorithm for part selection, sharing, and image categorization based on boosting. It embeds all three part types, without the need for explicit part type definition};
(iii) we introduce a fusion technique for combining part-based with global image representations. We report results competitive to the state-of-the-art on object, scene, action, and fine-grained categorization challenges, further improving over the very deep convolutional neural networks of~\cite{simonyan15,szegedy14}.

The rest of the paper is outlined as follows. In section 2, we describe related work, while section 3 outlines our proposed method. This is followed by the experimental evaluation in section 4. We draw conclusions in section 5.

\section{Related work}
\label{sec:relwork}

The state-of-the-art in image categorization relies on deep convolutional neural networks (ConvNets)~\cite{krizhevsky12,simonyan15,szegedy14,zeiler14}. These networks learn image feature representations in the form of convolution filters. Locality is implicitly incorporated by stacking and pooling local filter responses, resulting in increasingly larger filter responses, cumulating in a single global image representation. Two recent network architectures, the VGG network of Simonyan et al.~\cite{simonyan15} and the GoogleNet network of Szegedy et al.~\cite{szegedy14}, have shown that further increasing the network depth results in state-of-the-art performance on a wide range of image categorization datasets~\cite{simonyan15,szegedy14}.
A global representation from such networks benefits from augmenting it by aggregating local ConvNets features of densely sampled image parts in a VLAD~\cite{gong14} or Fisher Vector~\cite{cimpoi15,yoo2015multi} representation. We follow such approaches and augment the global representation by using ConvNets to represent discriminative local parts. 

An excellent source of part-based information for classifying object images is the response of a detector for the object at hand. In the pioneering work of Harzallah et al.~\cite{harzallah09}, it is shown that an object detector improves image classification, and vice versa~\cite{divvala2009empirical,sadeghi2011recognition}. Others improved upon this idea by adding context~\cite{song11}, and deep learning~\cite{oquab14}. However, when no bounding box annotations are available, a supervised detector cannot be trained. In our method, we do not use any bounding box annotations and rely exclusively on the global image label.

In the absence of bounding box annotations, one may focus on automatically discovering discriminative parts in images. The work of Singh et al.~\cite{singh12} proposes an unsupervised method for finding parts, by iterative sampling and clustering large collections of HOG features, SVM training on the clusters, and assigning new top members to each cluster.
Other part-based methods follow a supervised approach, e.g. using discriminative mode seeking~\cite{doersch13}, random forests~\cite{bossard14}, average max precision gains~\cite{endres13}, or group sparsity~\cite{sun13} for discovering the best parts for each category separately from image-level labels. Recent work moves away from HOG features in favor of local ConvNet activations for part selection~\cite{li2015mid}, based on association rule mining per category. In this work, we also leverage image-level labels for part selection. In contrast to~\cite{bossard14,doersch13,doersch12}, we also perform part selection by sharing over all categories, rather than performing selection per category. Furthermore, we optimize the part selection with the global image representation during fusion.

{\color{black}
The work by Juneja et al.~\cite{juneja13} shows that using image-level labels leads to better categorization performance using fewer parts than the unsupervised method of Singh et al.~\cite{singh12}. Their method selects parts that have the lowest entropy among categories. In effect, this limits the sharing over a few categories only. We strive to share parts as much as possible. Rather than relying on entropy for the selection, we prefer a boosting objective that optimizes sharing for all categories simultaneously.}

The seminal work of Torralba et al.~\cite{torralba07} has previously explored the idea of sharing among categories. We follow this line of work in a part-based setting. Similar to Torralba et al.~\cite{torralba07}, we opt for a boosting approach,
as we can exploit the inherent feature selection and sampling methods in AdaBoost~\cite{freund95} for jointly finding the best parts while training the category classifiers.
However, where Torralba et al.~\cite{torralba07} learn which parts distinguish multiple categories simultaneously from a common background, our objective is to learn what parts to share to distinguish categories from each other.
We extend boosting to explicitly focus on part sharing, and propose a bootstrapping method for fusion with the global image representation.

{\color{black}The work of Azizpour et al.~\cite{azizpour2015spotlight} generalizes latent variable models and shows how explicitly incorporating both Own (referred to as foreground) and Other (referred to as background) parts benefits objects in a binary setting.
We find empirically that Own and Other parts are indeed important, in addition to parts from Context. 
Furthermore, we extend the scope from a binary setting to both a multi-class and multi-label setting.
And finally, we introduce a method for fusion that exploits the strength of both global and part-based image representations.
}

{\color{black}Recent work of Parizi et al.~\cite{parizi15} opts for a joint optimization of part detection and image categorization, where the part detectors and image classifiers are optimized alternatively. They show that such a joint optimization over all categories simultaneously improves categorization performance over independently optimized part-based methods, be it that their approach requires significant feature dimension reduction to be feasible. Similar to~\cite{azizpour2015spotlight}, Parizi et al.~\cite{parizi15} consider Own and Other part types, where we establish the relevancy of three part types (Own, Other, and Context) for sharing among image categories and we demonstrate their effectiveness. Moreover, we use state-of-the-art ConvNet features without the need for dimension reduction. Finally, our algorithm naturally incorporates fusion with the global image representation.}

​
In~\cite{song2013discriminatively}, Song et al. learn tree classifiers on image parts for object detection. We also rely on trees, be it that we prefer to have many of them (order of $10^3$), and next to objects, we also consider scenes, actions, and fine-­grained birds.

\section{Part sharing for image categorization}
\label{sec:sharing}

\begin{figure*}[t]
\centering
\includegraphics[width=0.95\textwidth]{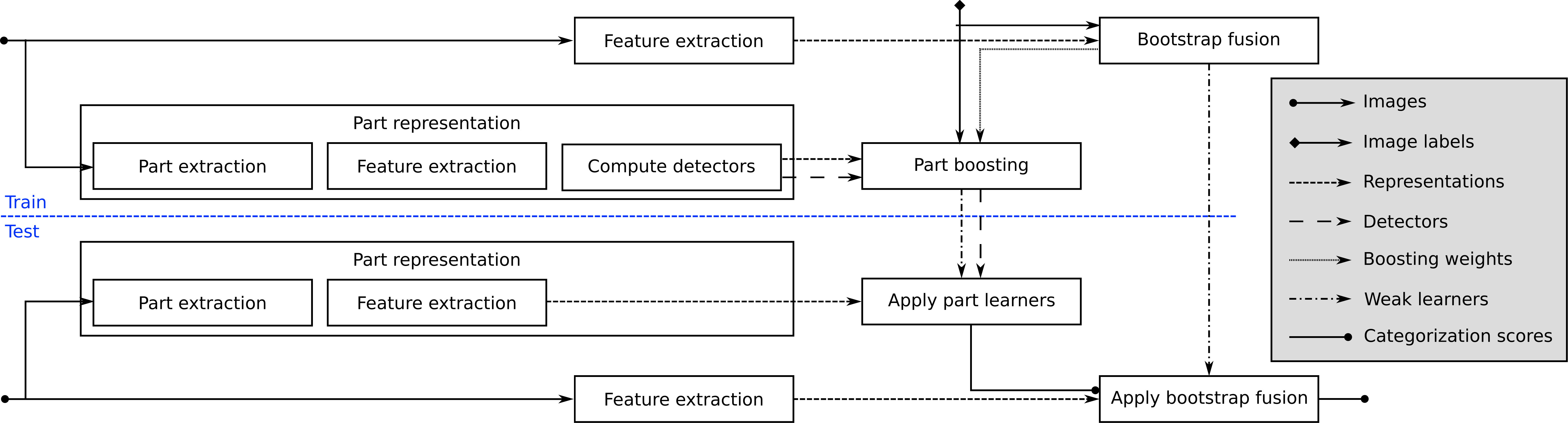}
\caption{\color{black}Overview of our part-based image representation with fusion. First, parts and their features are extracted. Second, we perform our part optimization. Third, we perform bootstrap fusion between the part-based and global image representations.}
\label{fig:overview}
\end{figure*}

In Figure~\ref{fig:overview}, we provide an overview of our approach. First, we perform part representation, where parts with their features and learned detectors are extracted from training images. Second, we perform our part-based optimization based on boosting to yield a part-based classifier. Third, we perform our bootstrap fusion between the part-based classifier and a classifier trained on global image representations.

\subsection{Part representation}
We decompose each image it into part proposals which offer a modest set of bounding boxes with a high likelihood to contain any part or object.
More formally, let the function $B(X)$ map an image $X$ to $k$ part proposals, $B(X) \rightarrow \{ p_1, p_2, \dots, p_k\}$.
{\color{black}For $n$ training images, $T = \{ X_1, X_2, \dots, X_n\}$, we extract and combine all parts as $P_{\text{train}} = \cup_{X \in T} B(X)$.}
In our work, we opt for selective search~\cite{uijlings13}, but our method applies to any part partition~\cite{cheng2014bing,krahenbuhl2014geodesic,zitnick2014edge}.

{\color{black}For each part $p_i$, we learn a part detector $d(\psi(p_i))$, where $\psi(p_i) \in \mathbb{R}^{d}$ denotes the feature representation of $p_i$.}
{\color{black}For the detector $d()$, we follow~\cite{juneja13} and transform the part representations into linear classifiers. Each linear classifier serves as an exemplar SVM for the corresponding part representation. We use the fast exemplar SVM approach of Hariharan et al.~\cite{hariharan12} to compute the classifiers. Let $\mu$ and $\Sigma^{1}$ denote the mean and covariance estimated from sampled part representation, ignoring their category label. Then the detector function is defined as: $d(p) = \Sigma^{-1} (p - \mu)$.}

Given a detector for each part, we represent an image $X$ by a feature vector $\mathbf{v}_X \in \mathbb{R}^{|P|}$, where each dimension corresponds to the response of a part detector on $X$. Each dimension $j$ in our image representation vector $\mathbf{v}^{(j)}_X$ is the best part-detector response for a training part $p_j$ in image $X$, 
\begin{equation}
\mathbf{v}_{X}^{(j)} = \max_{p \in B(x)} d(\psi(p_j)) \cdot \psi(p).
\end{equation}
Our image representation has $s=|P|$ dimensions where each dimension is a part from the set $P \subset P_{\text{train}}$. Our goal is to select the set of shared parts $P$ to use as detectors over all categories.

\begin{figure*}[t]
\centering
\includegraphics[width=0.875\textwidth]{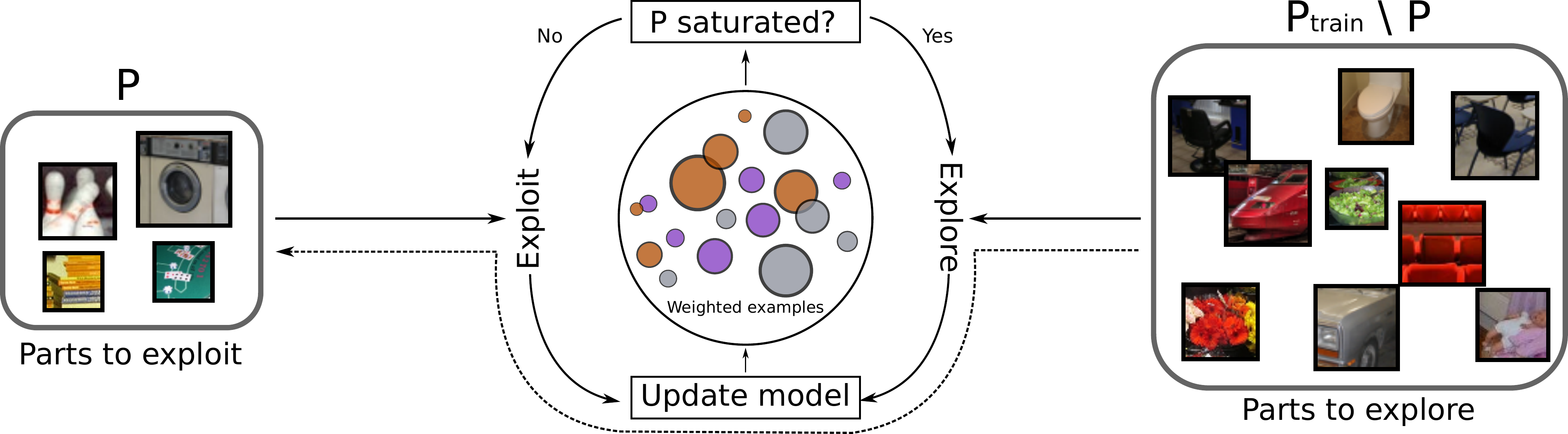}
\caption{Graphical overview of maximum exploiting sampling. The method starts by exploring a number of parts from the set $P_{\text{train}}$. After that, the selected parts are maximally exploited until their training performance saturates. This is then repeated multiple times by a new exploration step and a range of exploitation steps. In the figure, the colors of the training images represent different categories and the sizes represent the boosting weights $w_i$.}
\label{fig:statemachine}
\end{figure*}

\subsection{Part boosting}
We unify part selection $P \subset P_{\text{train}}$ and image categorization by extending Adaboost~\cite{freund95}. {\color{black}Boosting minimizes the training error in an iterative manner, converging asymptotically to the global minimum exponential loss~\cite{freund95}}. At each iteration $t$, a weak learner $f_{t}(\cdot;\Phi_t)$ selects the parts $\Phi_t$ and weights $\alpha_t$ that maximally decrease the current weighted image classification error while at the same time optimizing the part selection $P$.
{\color{black}The classification of a test image $\mathbf{v}_X$ is a weighted sum of the $T$ weak learners under the global constraint of using at most $s$ parts.}
\begin{equation}
h^{l}(\mathbf{X}) = \sum_{t=1}^{T} \alpha_{t} \cdot f_{t}(\mathbf{v}_X; \Phi_t), \quad \text{s.t.} ~| P| \leq s,
\label{eq:boost2}
\end{equation}
where $P = \cup_{t=1}^{T} \Phi_{t}$ is the set of all selected parts.
{\color{black}Here, the set $\Phi_t$ denotes the parts used to construct a weak learner at iteration $t$.}
Our formulation extends standard boosting with a global constraint $|P| \leq s$ on the number of parts.
{\color{black}By limiting the number of parts, we aim to arrive at a discriminative part-based representation~\cite{bossard14,doersch13,juneja13}. The convergence is dependent on the value of $s$; the higher the value, the more training iterations are required.}

{\color{black}In the objective, the weak learners $f_t (t=1,..,T)$ take the form of orthogonal decision trees~\cite{dietterich2000experimental}. The decision trees are constructed by selecting the part with the lowest weighted classification error at each binary split of the tree~\cite{freund95}.}
{\color{black}The weights of each weak learner $\alpha_t$ are computed as $\alpha_t = \frac{1}{2} \ln((1 - \epsilon_t) / \epsilon_t)$, where $\epsilon_t$ denotes the error rate of the weak learner~\cite{freund95}.}

For all $L$ categories we have $L$ corresponding binary classifiers of Eq.~\ref{eq:boost2}.
In a multi-label setting each classifier is independent; yet the set of parts $P$ are shared over all categories. In a multi-class setting we have to make a single choice per image:
\begin{equation}
h(\mathbf{X}) = \argmax_{l \in L} \frac{\sum_{t=1}^{T} \alpha_{t}^{l} \cdot f_{t}^{l}(\mathbf{v_X};\Phi_{t}^{l})}{\sum_{t=1}^{T} \alpha_{t}^{l}}, \quad \text{s.t.} | P | \leq s,
\label{eq:boost3}
\end{equation}
with now $P = \cup_{t=1}^{T} \cup_{l=1}^{L} \Phi_{t}^{l}$. Note that our added constraint  $| P| \leq s$ in Eq.~\ref{eq:boost3} states that at most $s$ unique parts over \emph{all} categories are used, enforcing the sharing of parts between categories.
{\color{black}By limiting the number of parts over all categories simultaneously, each part selected during representation learning should be discriminative for as many categories as possible, i.e. should be shared between categories.}
{\color{black} We opt for a one-vs-rest objective instead of a multi-class objective~\cite{eibl2005multiclass,saberian2011multiclass} to allow for both multi-class and multi-label image categorization.}

\textbf{Maximum exploiting sampling.}
We introduce a part sampling algorithm for re-using parts and sharing parts between categories, called \textit{maximum exploiting}. The main idea of maximum exploiting is to gradually increase the cardinality of $P$ by balancing exploration: selection of new parts from $P_{\text{train}}$; and exploitation: reusing and sharing of previously selected parts $P$. 

{\color{black}The idea behind Maximum Exploit sampling builds upon the work by Freund and Shapire~\cite{freund95}, which states that the training error on a strong learner reduces exponentially with the number of non-random weak learners. Here, we use this result for joint part selection and image categorization. As long as we can train non-random weak learners on the same set of parts, we are guaranteed to reduce the training error rate. We exploit this maximally in our sampling approach.}

At train iteration $t$, we force the decision tree $f_{t}(\cdot;\Phi_{t})$ to select parts $\Phi_{t}$ exclusively from $P$, exploiting the shared pool. When the classification performance saturates and $P$ is maximally exploited, we allow a single exploration step, selecting new parts $\Phi_{t}$ from $P_{\text{train}} \setminus P$. In Fig.~\ref{fig:statemachine} we illustrate our algorithm. Our sampling scheme minimizes the cardinality of the selected parts $P$, forcing the decision trees to reuse the shared parts from all categories as long as this positively affects the empirical loss over the training examples.

To determine if $P$ is maximally exploited, we measure the saturation of $P$ by the classification error within the last range of exploitation iterations. For a category $l$, let this range start at iteration $u$ and let the current iteration be $v$. Then the error is defined as:
\begin{equation}
\epsilon_{x,y}^{l} = \sum_{i=1}^{N} \bigg[\text{sign}(\sum_{t=u}^{v} \alpha_{t}^{l} \cdot f_{t}^{l}(\mathbf{v}_{X_i};\Phi_{t}^{l})) \not= Y_{i}^{l} \bigg],
\label{eq:maxexp}
\end{equation}
where $Y_{i}^{l} \in \{-1, +1\}$ states whether image $i$ is of category $l$. The error of Eq.~\ref{eq:maxexp} states the number of miss-classifications using the weak classifiers created in the current exploitation iterations. Let $\epsilon_{u,v}$ denote the average error over all categories within range $[u,v]$, i.e.:
\begin{equation}
\epsilon_{u,v} = \sum_{l=1}^{L} \frac{\epsilon_{u,v}^{l}}{L}.
\end{equation}
We keep exploiting if $\epsilon_{u,v} < \epsilon_{u,v-1}$, otherwise, we perform a single exploration step and restart our exploitation evaluation, setting $u=v$.

\begin{figure}[t]
\centering
\includegraphics[width=0.6\linewidth]{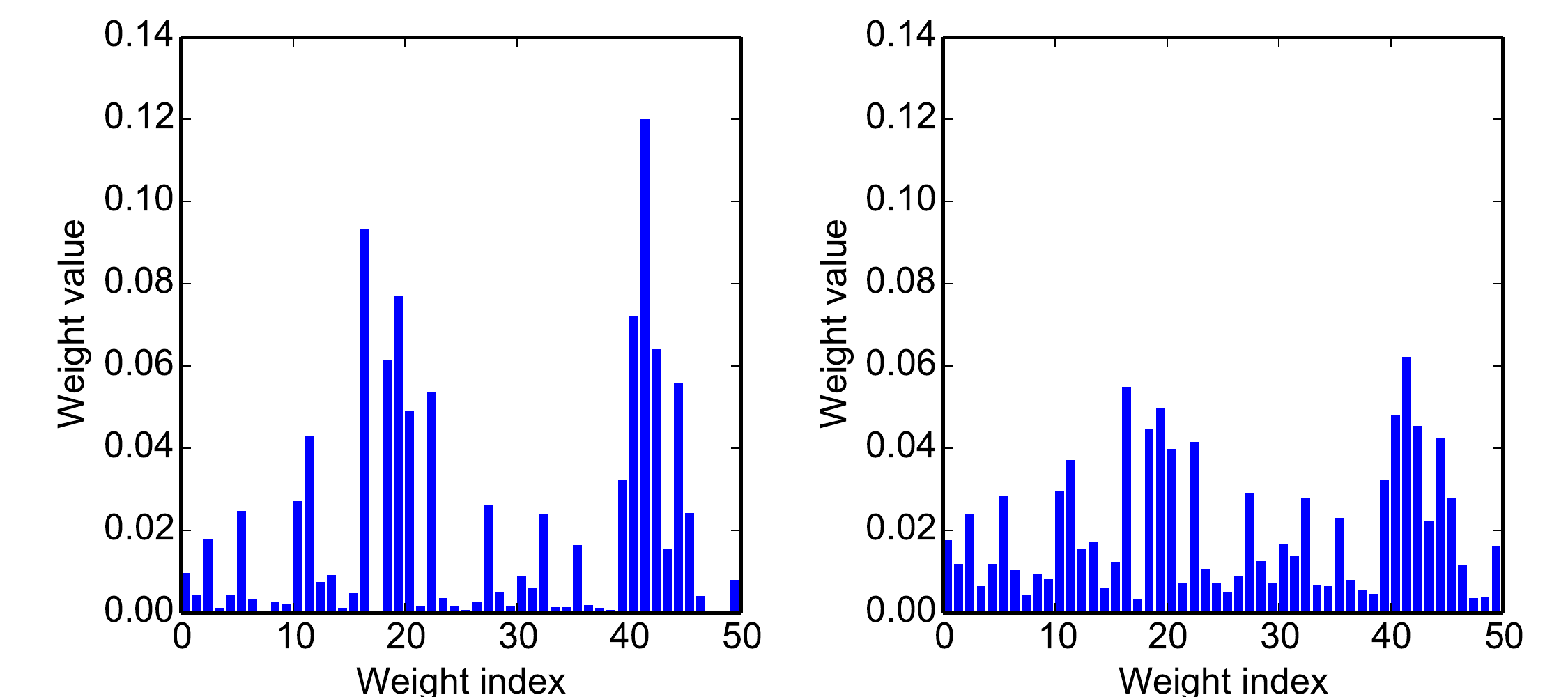}
\caption{The effect of power normalization for a synthetic set of weights from 50 training examples. On the left we show the weights before power normalization, on the right after normalization. By re-balancing the weights of the training examples, we focus on the hard examples without disregarding the other examples.}
\label{fig:powernorm}
\end{figure}

\subsection{Bootstrap fusion}
We extend our shared part selection and sampling to a fusion with the global image representation. Instead of early or late fusion that would combine the part-based representation independently with the global image representation~\cite{doersch13,juneja13,snoek05}, we fuse the representations from the start. We propose \emph{bootstrap fusion}, which jointly optimizes part-selection and image classification by selecting those parts that gain the most from fusing with the global image representation.

{\color{black} In our bootstrap fusion, we start by applying our maximum exploitation on the global image representation F. Afterwards, our idea is to bootstrap the AdaBoost weights of the training images as initialization for the part representations. This bootstrapping provides information about the difficulty per training image for categorization according to the global representation. By transferring the weights to the part-based model, the part-based representation learning is forced to focus on the mistakes caused by the global representation. This fusion procedure enhances the complementary nature between the part and global representations. In our fusion approach, the dimensionality is equal to the dimensionality of the global representation and the number of selected parts $s$ in the part representation.}

When transferring the weights, certain training examples have weight values $w_{i}$ up to several orders of magnitude higher or lower than the average weight value. The reason for this is the presence of the exponent in the weight update in AdaBoost~\cite{freund95}. Training examples that are consistently correctly or incorrectly classified in the initial boosted classifier will yield extreme weight values. Although we want to focus on the hard examples, we are not interested in focusing solely on these examples. To combat the high variance of the weights, we perform a power normalization step~\cite{sanchez13}, followed by an $\ell_{1}$ normalization to make the weights a distribution again:
\begin{equation}
w_{i} = \frac{w_{i}^{\alpha}}{\sum_{j=1}^{N} w_{j}^{\alpha}}.
\end{equation}
Throughout our experiments, we set $\alpha = \frac{1}{2}$. The use of power normalization results in a more balanced weight distribution, which in turn results in better fusion performance. Fig.~\ref{fig:powernorm} highlights the effect of power normalization on the weight values.

During testing, we apply the same part and feature extraction as during training. We apply our boosted classifier with the $s$ selected parts to the parts of the test image. Idem, the boosted global classifier is applied to the global image representation of the test image. The confidence scores of both classifiers are summed and the category with the highest confidence is selected as the target class.

\section{Experiments}
\label{sec:exp}

\subsection{Datasets}
In our experiments, we consider 4 datasets.

\textbf{Pascal VOC 2007}~\cite{everingham10} consists of 9,963 images and 20 object categories, where we report on the provided trainval/test split using the Average Precision score~\cite{everingham10}.

\textbf{MIT 67 Indoor Scenes}~\cite{quattoni09} consists of 6,700 images and 67 scene categories, where we use the provided 80/20 split for each scene category~\cite{quattoni09}, reporting results using multi-class classification accuracy.

\textbf{Willow Action}~\cite{delaitre10} consists of 911 images and 7 action categories, where we use the train/test split of~\cite{delaitre10}, reporting with the Average Precision score.

{\color{black}\textbf{Calltach-UCSD Birds 200-2011}~\cite{wah2011caltech} consists of 11,788 images and 200 fine-grained bird categories, where we use the train/test split of~\cite{wah2011caltech}, reporting with the multi-class classification accuracy.}

\subsection{Implementation details}

\emph{Part extraction.} {\color{black}During both training and testing, we use selective search as the function $B(X)$ which splits each image into roughly 2,000 parts~\cite{uijlings13}.
Due to the hierarchical nature of selective search, we observe it in fact generates parts with varying sizes, from superpixels to the complete image. Although selective search is intended for objects, we observe it in fact generates parts suitable for our purpose. On Pascal VOC 2007, we derived there are over 22 parts for each labeled object for which the \emph{part of} score from Vezhnevets and Ferrari~\cite{vezhnevets15} is at least 0.5.}

\emph{Feature extraction.} As features, we employ a GoogLeNet convolutional neural network~\cite{szegedy14}, pre-trained on 15k ImageNet categories~\cite{imagenet09}. For a given part, we rescale it to 256$\times$256 pixels and feed it to the network, resulting in a 1,024-dimensional feature vector, which is subsequently $\ell_{2}$ normalized.

\emph{Part detection.} {\color{black}We use the fast exemplar SVM of Hariharan et al.~\cite{hariharan12} to transform the part representations into linear classifiers. The linear classifier of a part is used for the max-pooling operation of Eq. 1.}
{\color{black}For fair comparison, the max-pooling operation is only applied to the whole image. Both~\cite{juneja13} and our approach will improve further when we incorporate pooling over the scene layout by~\cite{lazebnik2006beyond}.}
We sample roughly 400k part features from train images to estimate the corresponding $\mu$ and $\Sigma$ parameters.

\emph{Part boosting.} {\color{black} For the weak learners, we opt for orthogonal decision trees, as AdaBoost with decision trees have shown to yield excellent categorization performance~\cite{dietterich2000experimental}.} At each split, we select the part that minimizes the normalized weighted miss-classification error of the examples in the corresponding node.
{\color{black} As shown by B\"{u}hlmann and Yu~\cite{buhlmann2003boosting}, AdaBoost is resistant to overfitting when using a large number of training iterations. Therefore, we use 2,000 iterations for boosting throughout our experiments for training the image classifiers.}

\begin{figure*}[t]
\centering
\includegraphics[width=\textwidth]{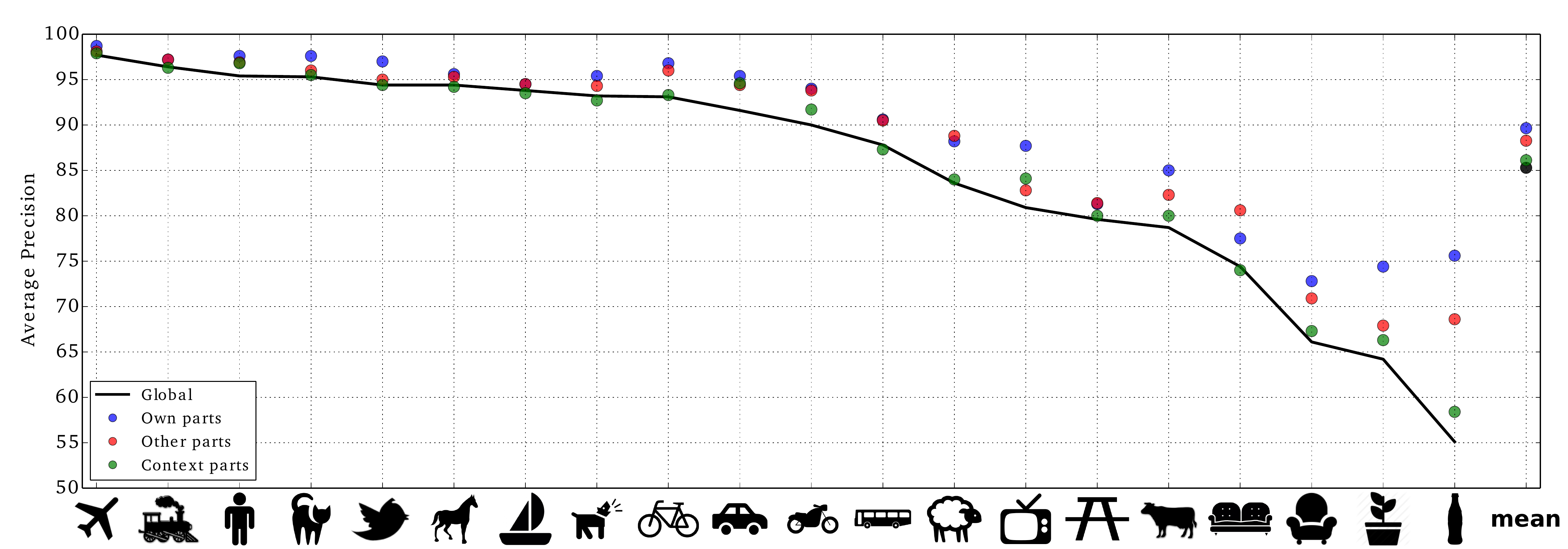}
\caption{Average Precision (\%) on Pascal VOC 2007 for the global image representation~\cite{szegedy14} and its combination with each type of category parts (approximated from provided bounding boxes). We conclude that all types of category parts matter.}
\label{fig:exp1-res}
\end{figure*}

\subsection{Experimental evaluation}

\begin{figure*}[t]
\centering
\begin{subfigure}[b]{0.225\linewidth}
Own category parts.
\centering
\includegraphics[width=\linewidth]{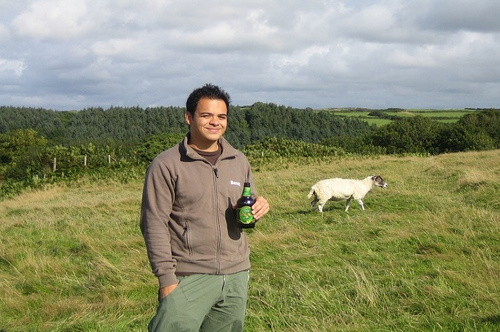}
\caption{\includegraphics[height=0.0125\textheight]{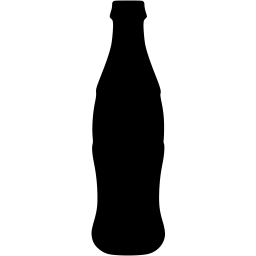} 4097 $\rightarrow$ 340}
\label{fig:qual-exp1-a}
\end{subfigure}
\begin{subfigure}[b]{0.225\linewidth}
\centering
\includegraphics[width=\linewidth]{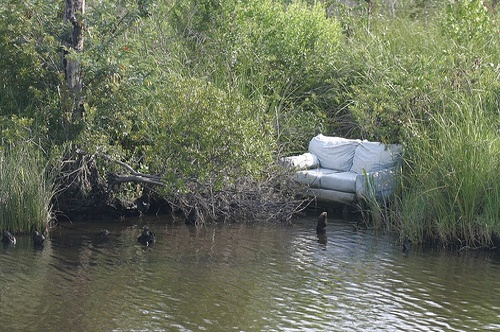}
\caption{\includegraphics[height=0.0125\textheight]{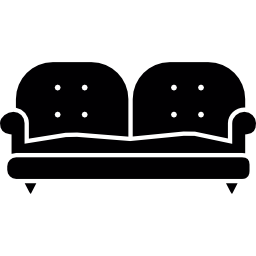} 4446 $\rightarrow$ 2081}
\label{fig:qual-exp1-b}
\end{subfigure}
\begin{subfigure}[b]{0.225\linewidth}
\centering
\includegraphics[width=\linewidth]{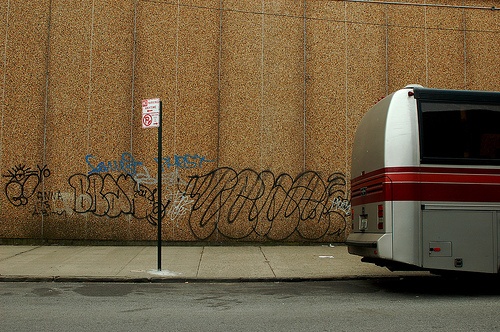}
\caption{\includegraphics[height=0.0125\textheight]{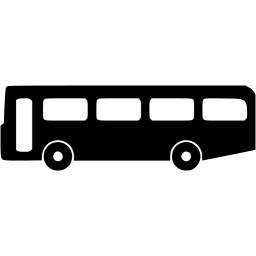} 262 $\rightarrow$ 130}
\label{fig:qual-exp1-c}
\end{subfigure}
\hspace{0.05cm} \unskip\ \vrule\ \hspace{0.05cm}
\begin{subfigure}[b]{0.225\linewidth}
\centering
\includegraphics[width=\linewidth]{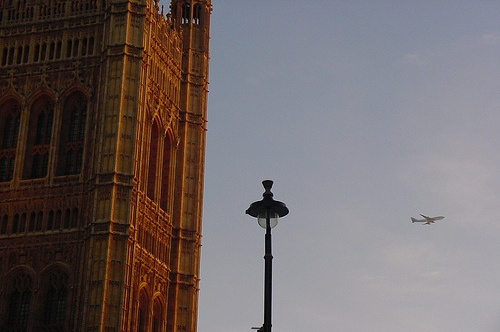}
\caption{\includegraphics[height=0.0125\textheight]{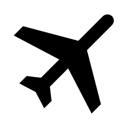} 456 $\rightarrow$ 539}
\label{fig:qual-exp1-d}
\end{subfigure}
\vspace{0.35cm}\\
\begin{subfigure}[b]{0.225\linewidth}
Other category parts.
\centering
\includegraphics[width=\linewidth]{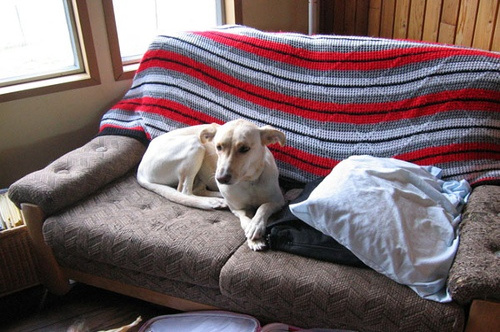}
\caption{\includegraphics[height=0.0125\textheight]{images/icons/sofa.png} 372 $\rightarrow$ 128}
\label{fig:qual-exp1-e}
\end{subfigure}
\begin{subfigure}[b]{0.225\linewidth}
\centering
\includegraphics[width=\linewidth]{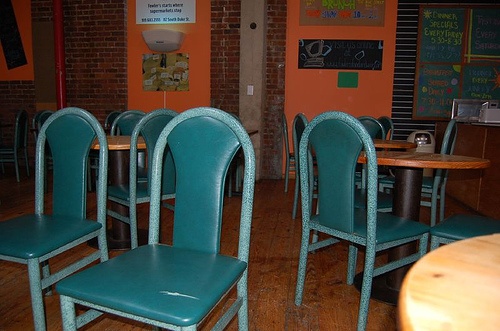}
\caption{\includegraphics[height=0.0125\textheight]{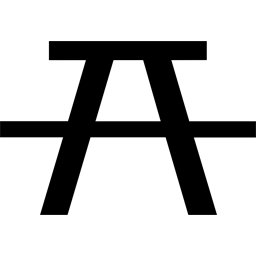} 1443 $\rightarrow$ 404}
\label{fig:qual-exp1-f}
\end{subfigure}
\begin{subfigure}[b]{0.225\linewidth}
\centering
\includegraphics[width=\linewidth]{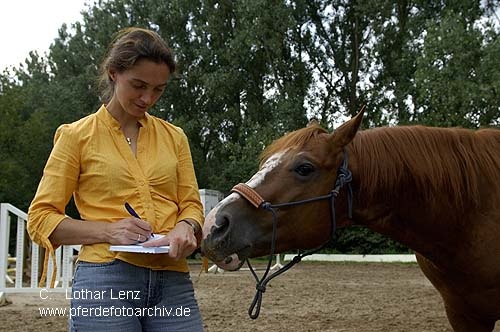}
\caption{\includegraphics[height=0.0125\textheight]{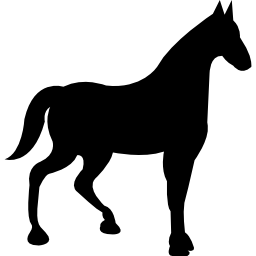} 120 $\rightarrow$ 80}
\label{fig:qual-exp1-g}
\end{subfigure}
\hspace{0.05cm} \unskip\ \vrule\ \hspace{0.05cm}
\begin{subfigure}[b]{0.225\linewidth}
\centering
\includegraphics[width=\linewidth]{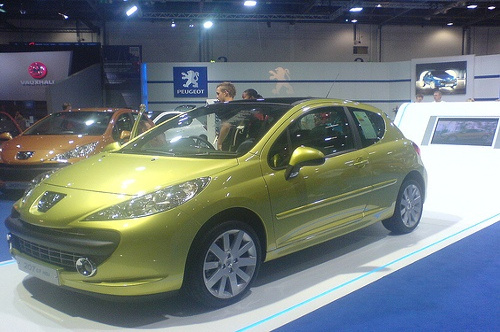}
\caption{\includegraphics[height=0.0125\textheight]{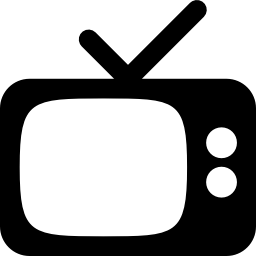} 3748 $\rightarrow$ 4527}
\label{fig:qual-exp1-h}
\end{subfigure}
\vspace{0.35cm}\\
\begin{subfigure}[b]{0.225\linewidth}
Context parts.
\centering
\includegraphics[width=\linewidth]{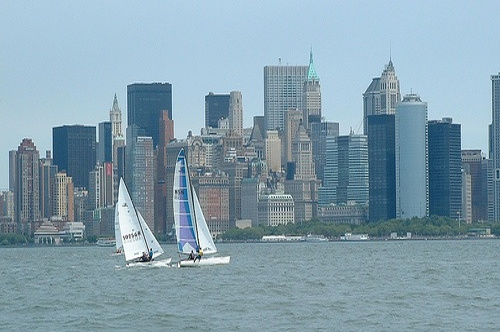}
\caption{\includegraphics[height=0.0125\textheight]{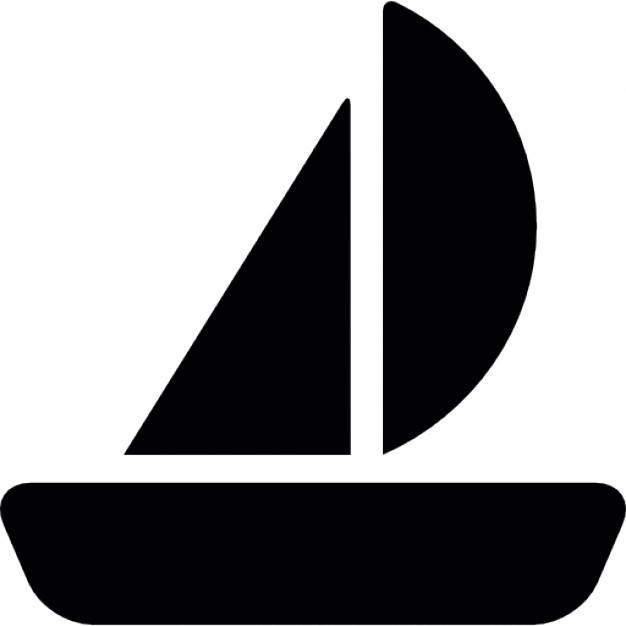} 30 $\rightarrow$ 12}
\label{fig:qual-exp1-i}
\end{subfigure}
\begin{subfigure}[b]{0.225\linewidth}
\centering
\includegraphics[width=\linewidth]{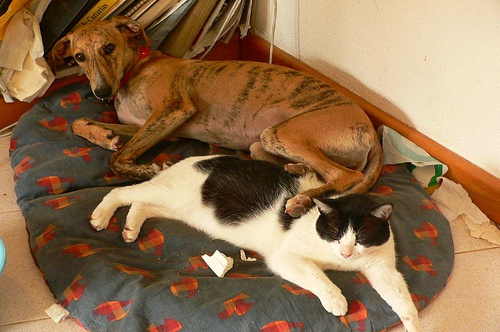}
\caption{\includegraphics[height=0.0125\textheight]{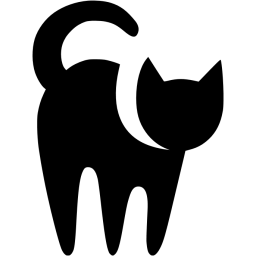} 436 $\rightarrow$ 309}
\label{fig:qual-exp1-j}
\end{subfigure}
\begin{subfigure}[b]{0.225\linewidth}
\centering
\includegraphics[width=\linewidth]{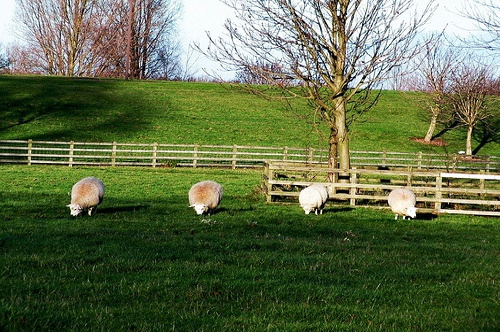}
\caption{\includegraphics[height=0.0125\textheight]{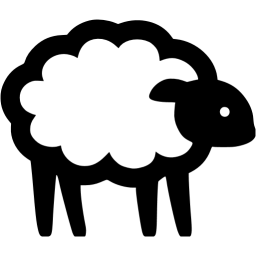} 796 $\rightarrow$ 242}
\label{fig:qual-exp1-k}
\end{subfigure}
\hspace{0.05cm} \unskip\ \vrule\ \hspace{0.05cm}
\begin{subfigure}[b]{0.225\linewidth}
\centering
\includegraphics[width=\linewidth]{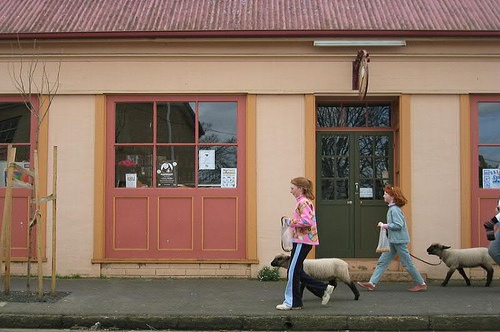}
\caption{\includegraphics[height=0.0125\textheight]{images/icons/sheep.png} 479 $\rightarrow$ 560}
\label{fig:qual-exp1-l}
\end{subfigure}
\caption{Qualitative examples of when the different part types improve/decrease the categorization performance. Shown are improvements using own (a-d), other (e-h), and context (i-l) parts. The first three examples have increased performance, the last example has decreased performance. The icon denotes the category, while the numbers state the original and new ranking position.}
\label{fig:qual-exp1}
\end{figure*}

\subsubsection*{Experiment 1: All part types matter}
We first motivate our main hypothesis that part-based methods benefit from three types of category parts. We rely on the Pascal VOC 2007 dataset for this experiment. As a surrogate for each type of part, we use the object bounding boxes as part locations. For each category, we use the features from the bounding boxes of the same category as its \emph{own} parts. Similarly, the features from the bounding boxes of all other categories are used as \emph{other} parts. For \emph{context} parts, we use the features from selective search boxes without overlap to any ground truth bounding box. As a baseline, we use the performance of the global image representation. We add the features of each part type to the global representation to evaluate their effect.
\\\\
\textbf{Results.} We show the image categorization results for the global representation and its combination with each part type in Figure~\ref{fig:exp1-res}. As the Figure indicates, adding each of the different part types boosts the categorization performance. Not surprisingly, the use of own parts for each category yields the best improvement (89.6\% vs. 85.3\% mean Average Precision (mAP)). We especially note the increase obtained for small objects such as \texttt{bottle} (from 55.1\% to 76.3\%) and \texttt{potted plant} (from 64.2\% to 74.4\%).
\begin{figure}[t]
\centering
\includegraphics[width=0.49\linewidth]{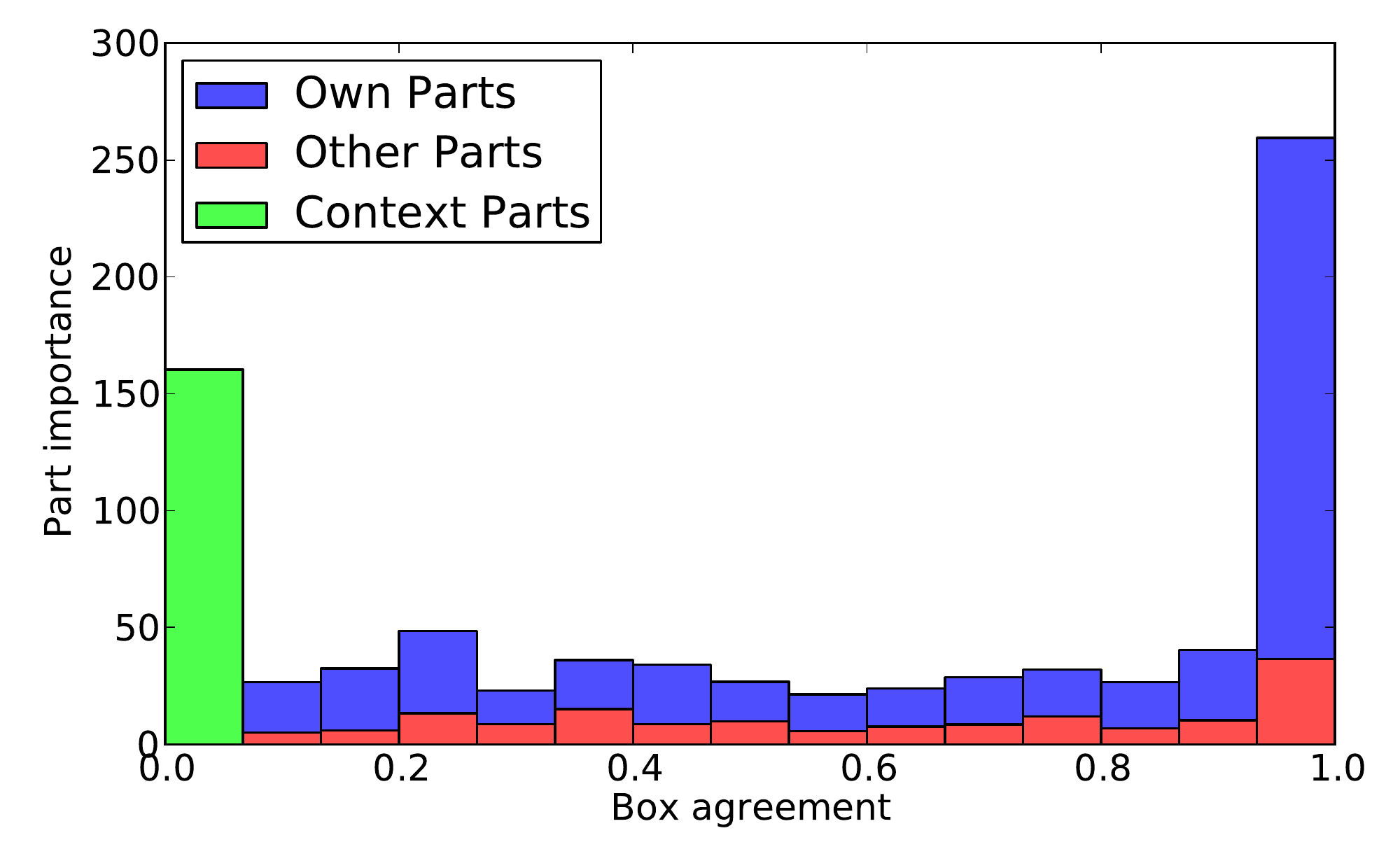}
\caption{Histogram showing the importance of our selected parts, as a function of the overlap with annotated objects. On the x-axis, we show the distribution of the overlap between our parts and ground truth objects. On the y-axis, we show how important the parts have been during training. The plot shows that our parts focus on parts from its own category (blue), while also taking in other categories (red) and context (green).}
\label{fig:exp2-1}
\end{figure}

\begin{figure*}[t]
\begin{subfigure}[b]{0.32\textwidth}
\centering
\includegraphics[width=0.49\linewidth]{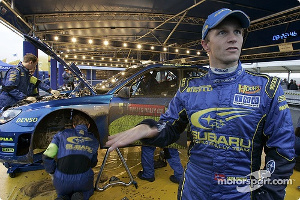}
\includegraphics[width=0.49\linewidth]{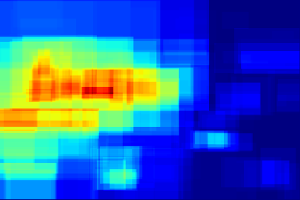}
\caption{Car.}
\label{fig:qual-exp2-a}
\end{subfigure}
\hspace{0.15cm}
\begin{subfigure}[b]{0.32\textwidth}
\centering
\includegraphics[width=0.49\linewidth]{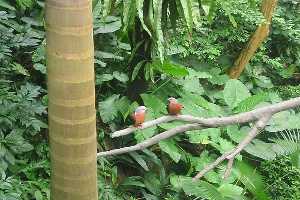}
\includegraphics[width=0.49\linewidth]{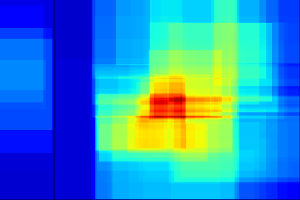}
\caption{Bird.}
\label{fig:qual-exp2-b}
\end{subfigure}
\hspace{0.15cm}
\begin{subfigure}[b]{0.32\textwidth}
\centering
\includegraphics[width=0.49\linewidth]{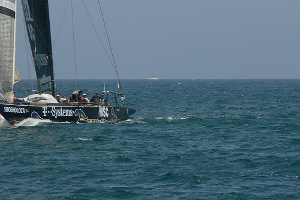}
\includegraphics[width=0.49\linewidth]{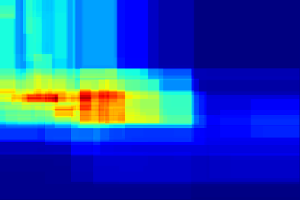}
\caption{Boat.}
\label{fig:qual-exp2-c}
\end{subfigure}
\begin{subfigure}[b]{0.32\textwidth}
\centering
\includegraphics[width=0.49\linewidth]{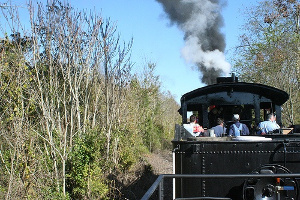}
\includegraphics[width=0.49\linewidth]{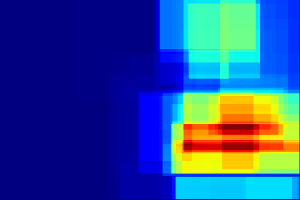}
\caption{Train.}
\label{fig:qual-exp2-d}
\end{subfigure}
\hspace{0.15cm}
\begin{subfigure}[b]{0.32\textwidth}
\centering
\includegraphics[width=0.49\linewidth]{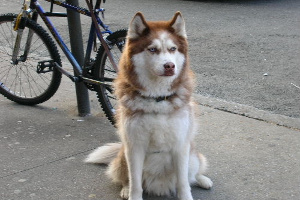}
\includegraphics[width=0.49\linewidth]{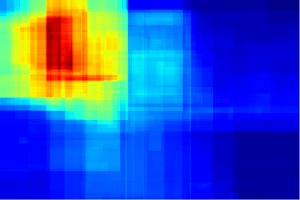}
\caption{Bike.}
\label{fig:qual-exp2-e}
\end{subfigure}
\hspace{0.15cm}
\begin{subfigure}[b]{0.32\textwidth}
\centering
\includegraphics[width=0.49\linewidth]{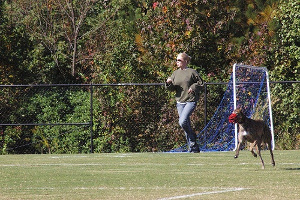}
\includegraphics[width=0.49\linewidth]{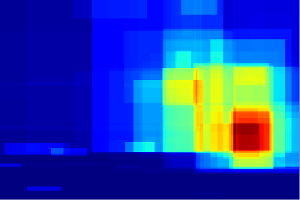}
\caption{Dog.}
\label{fig:qual-exp2-f}
\end{subfigure}
\begin{subfigure}[b]{0.32\textwidth}
\centering
\includegraphics[width=0.49\linewidth]{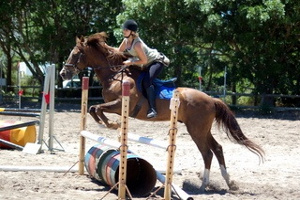}
\includegraphics[width=0.49\linewidth]{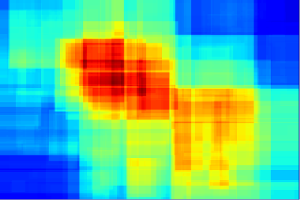}
\caption{Horse.}
\label{fig:qual-exp2-g}
\end{subfigure}
\hspace{0.15cm}
\begin{subfigure}[b]{0.32\textwidth}
\centering
\includegraphics[width=0.49\linewidth]{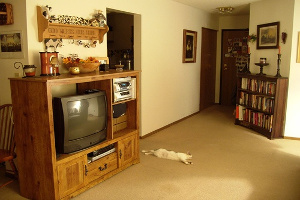}
\includegraphics[width=0.49\linewidth]{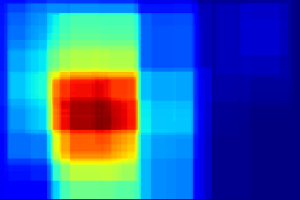}
\caption{Tv.}
\label{fig:qual-exp2-h}
\end{subfigure}
\hspace{0.15cm}
\begin{subfigure}[b]{0.32\textwidth}
\centering
\includegraphics[width=0.49\linewidth]{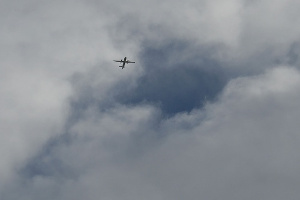}
\includegraphics[width=0.49\linewidth]{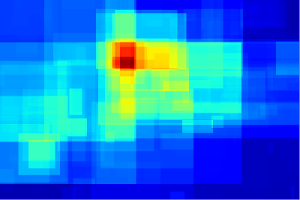}
\caption{Aeroplane.}
\label{fig:qual-exp2-i}
\end{subfigure}
\caption{Examples of the parts in the image where our method fires for nine object categories. Our parts focus on parts from the object itself, as well as other category parts and context. Note that our method automatically discovers when to share and not to share; dog parts are not useful for bike (e), but human parts are useful for dog (f).}
\label{fig:qual-exp2}
\end{figure*}

Surprisingly effective is the addition of other parts, with an mAP of 88.6\%. For multiple categories, such as \texttt{sofa}, it is even more favorable to add parts from other categories over parts from their own category. More specifically, using parts from \texttt{cats} and \texttt{dogs} yields a better performance than using parts from \texttt{sofa} itself. This result highlights the importance of exploiting and sharing the co-occurrence of categories. Lastly, the overall performance of context parts is marginally beneficial, compared to own and other parts (86.1\% mAP). For categories such as \texttt{car} (94.0\%), \texttt{motor bike} (91.7\%), and \texttt{tv} (84.1\%), the use of context parts may be good choice.

{\color{black}We note that experiment 1 serves to show that all part types matter for image categorization. Since we rely for this experiment on ground truth bounding boxes to approximate parts, the results should not directly be compared to other part-based methods, who all rely on the class labels only. A direct comparison will be discussed in experiment 2 and beyond.}
\\\\
\textbf{Qualitative analysis.} To understand why each part type matters for image categorization, we provide additional qualitative results. In Figure~\ref{fig:qual-exp1}, we show test examples that significantly improve or decrease in predictive performance when each of the part types is added.

For own parts, we observe that the improvement is most significant for small objects, objects out of context, and occluded objects, as shown in Figure~\ref{fig:qual-exp1-a}-\ref{fig:qual-exp1-c} for respectively \texttt{bottle}, \texttt{sofa}, and \texttt{bus}. Exploiting and sharing parts from other categories is beneficial when there is a high co-occurrence between categories in the images. This pattern is shown in Figure~\ref{fig:qual-exp1-e}-\ref{fig:qual-exp1-g}, for \texttt{sofa} / \texttt{dog}, \texttt{chair} / \texttt{dining table}, and \texttt{horse} / \texttt{person}. Parts from context are similarly beneficial in case of a co-occurrence between the categories and contextual elements. Notable examples are buildings for \texttt{boat}, cat basket for \texttt{cat}, and fence for \texttt{sheep}, as shown in Figure~\ref{fig:qual-exp1-i}-\ref{fig:qual-exp1-k}.

Figure~\ref{fig:qual-exp1} furthermore shows where each part type fails. For own category parts, performance decreases e.g. when the object can not be located, as shown in Figure~\ref{fig:qual-exp1-d}. Other category parts fail for inconsistent co-occurrence, such as \texttt{car} and \texttt{tv} in Figure~\ref{fig:qual-exp1-h}. For context parts, performance decreases when context is either not present or consistent with other object categories, as shown in Figure~\ref{fig:qual-exp1-l}.

Figure~\ref{fig:share-example} also shows that different part types focus on different elements in the scene, taking in the complete information from the image. Based on the quantitative and qualitative results of this experiment, we conclude that all part types matter for image categorization and should be exploited to improve image categorization.

\subsubsection*{Experiment 2: Evaluating our part-based method}
We evaluate our joint part selection and image categorization without box information in three steps: (i) evaluate whether we capture all part types, (ii) compare to separate part optimization, and (iii) compare our maximum exploit sampling to baseline samplers.
\\\\
\noindent
\textbf{Do we capture all types of parts?}
First, we validate that our method follows our hypothesis and we ask ourselves; do we capture all part types? We perform this evaluation on Pascal VOC 2007, as the object annotations of this dataset enables such an evaluation. To validate that our method is capable of incorporating all part types, we analyze the importance of each selected part as a function of the box agreement. We run our algorithm to select a total of 500 parts across all 20 categories in Pascal VOC 2007. We note that this setting yields an mAP of 89.1\%, significantly higher than the global representation and on par with the representations from the strongly supervised bounding boxes used in the first experiment. For each selected part $p$, we compute its importance in a single decision tree as the normalized miss-classification reduction; this value is summed over all the decision tree where the part is used. For the selected part $p$ and the best overlapping box $b$, the box agreement is computed here as $\frac{p \cap b}{p}$~\cite{vezhnevets15}. Intuitively, the box agreement states to what extend $p$ is part of $b$.

The relation between part importance and box agreement is shown in Figure~\ref{fig:exp2-1}. The Figure shows two clear peaks. The leftmost peak indicates that our method utilizes parts with no overlap to ground truth boxes, i.e. context parts. The other peak is at a box agreement of 1; when a part is contained in an object. The red and blue bars indicate that each category uses parts from its own object and from other objects. From the Figure we conclude that our method uses own (blue), other (red), and context (green) parts.

\begin{figure*}[t]
\begin{subfigure}[b]{0.65\textwidth}
\centering
\includegraphics[width=0.49\linewidth]{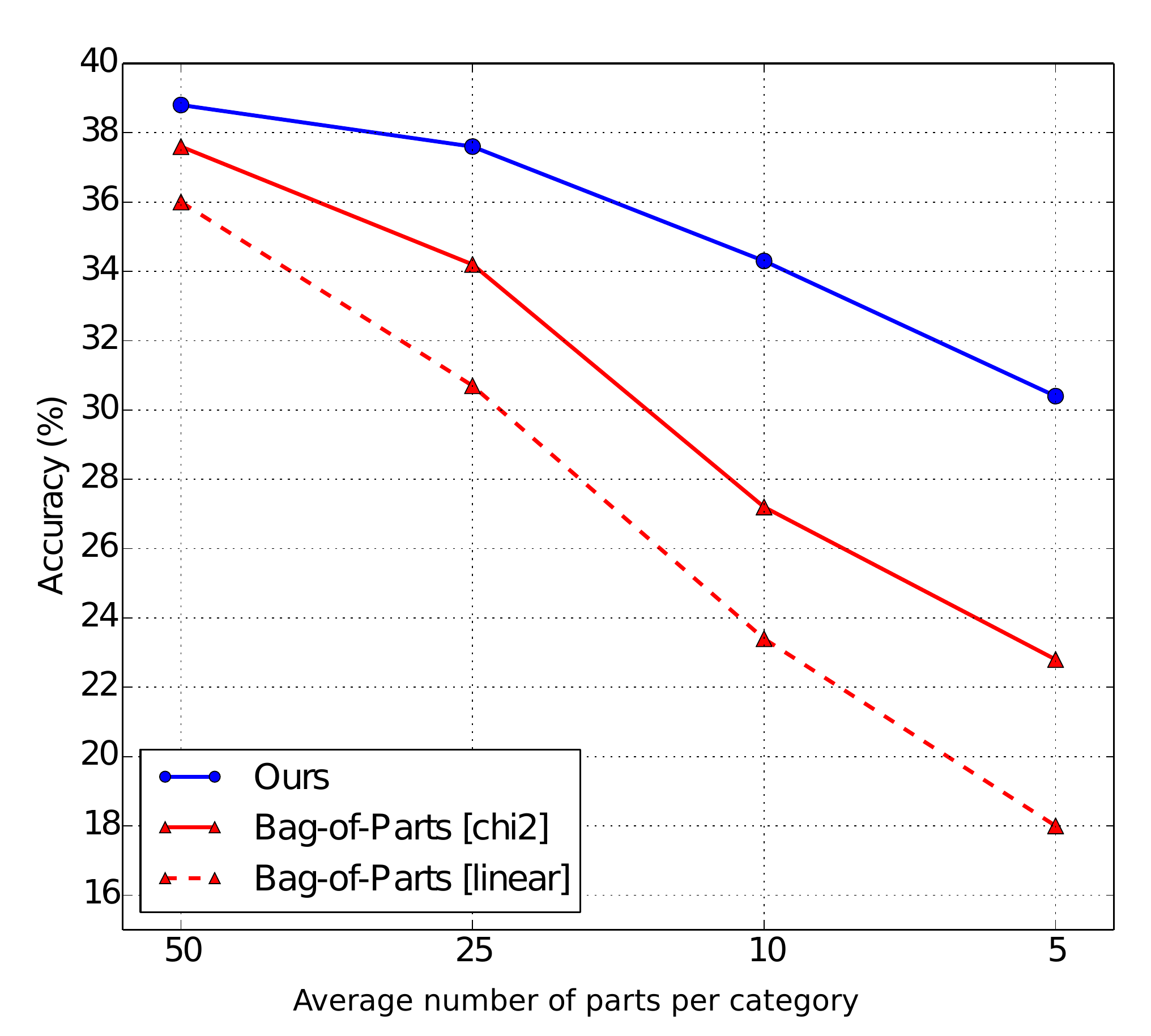}
\includegraphics[width=0.49\linewidth]{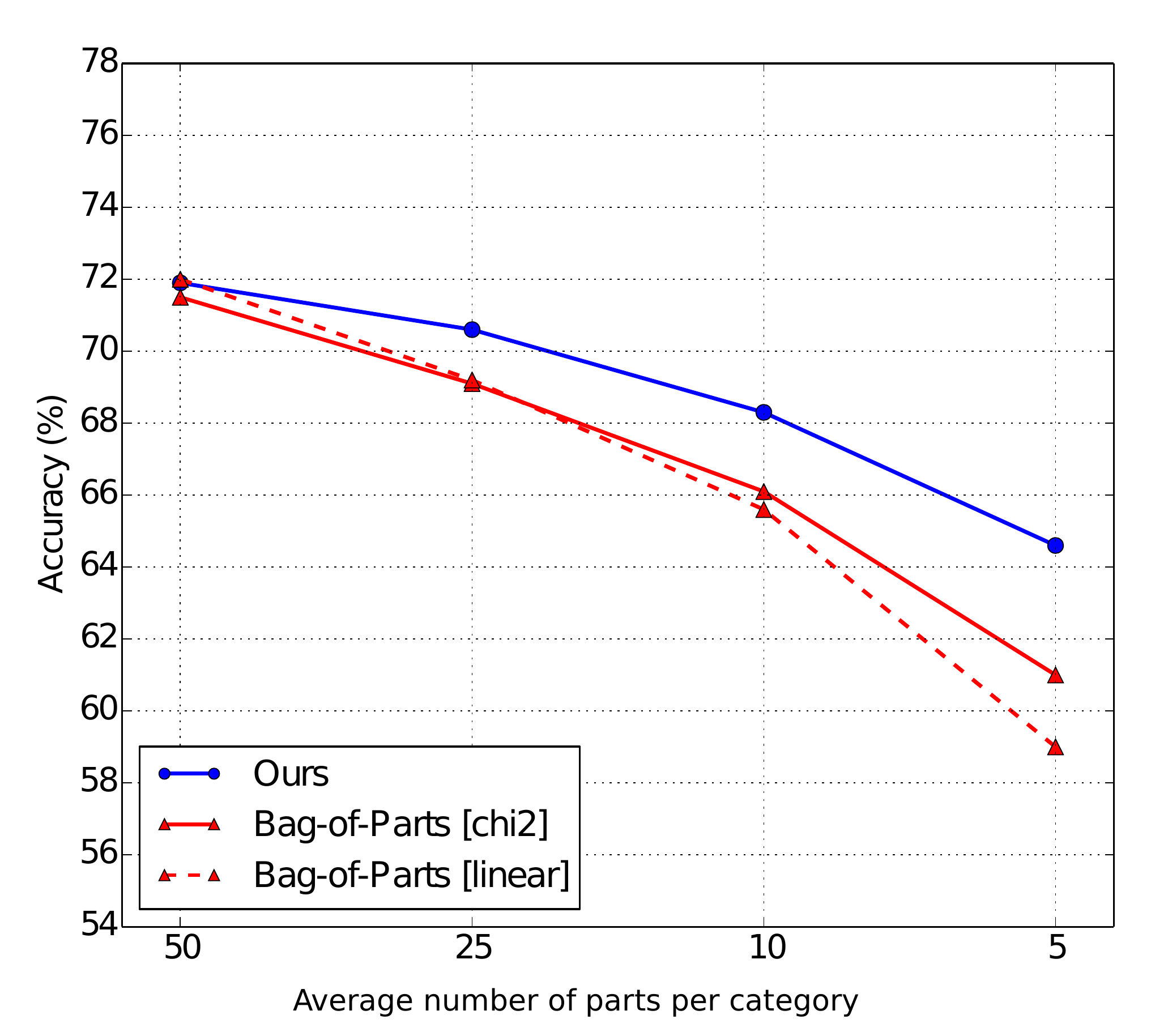}
\caption{Our method vs. the separate optimization of Juneja et al.~\cite{juneja13}.}
\label{fig:exp2-2-3-a}
\end{subfigure}
\begin{subfigure}[b]{0.32\textwidth}
\centering
\includegraphics[width=\linewidth]{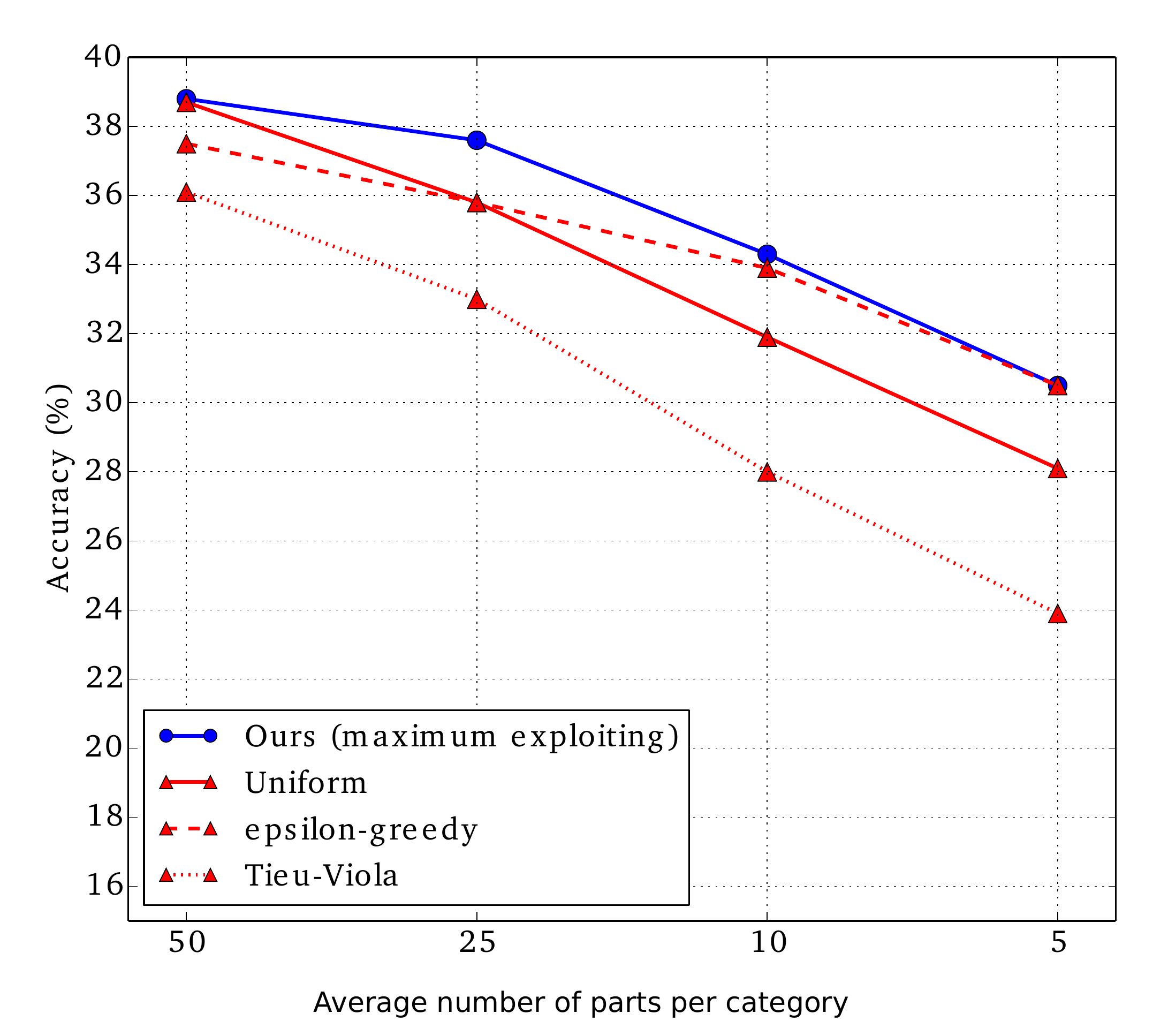}
\caption{Max. exploit vs. basic samplers.}
\label{fig:exp2-2-3-b}
\end{subfigure}
\caption{Classification results on MIT 67 Indoor Scenes. In \ref{fig:exp2-2-3-a}, we compare our method to the separate optimization of~\cite{juneja13}. In \ref{fig:exp2-2-3-b}, we compare the maximum exploit to baseline AdaBoost samplings. In both experiments, our approach is preferred over the baselines.}
\label{fig:exp2-2-3}
\end{figure*}

\begin{figure*}[t]
\begin{subfigure}[b]{0.33\textwidth}
\centering
\includegraphics[width=\linewidth]{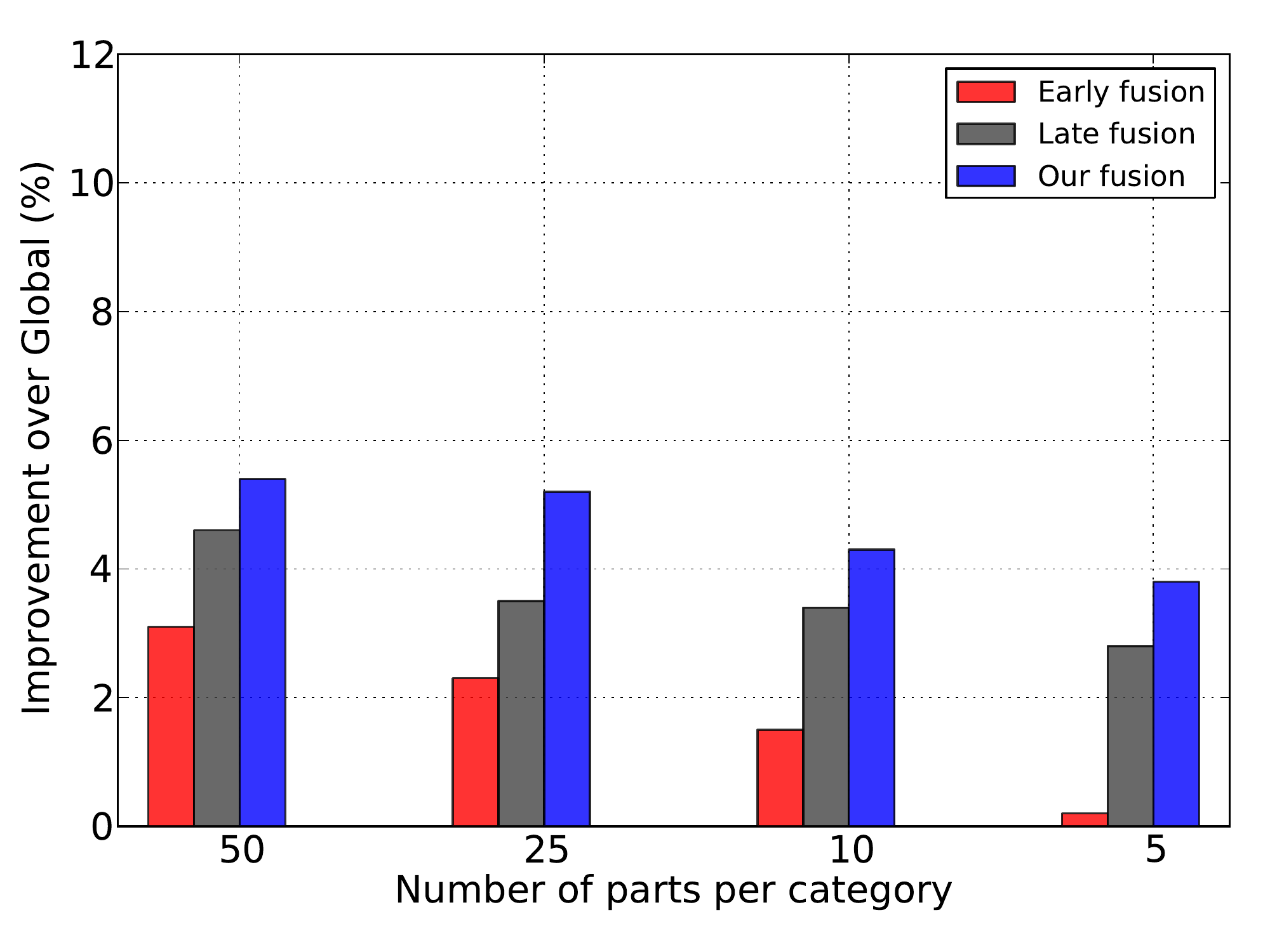}
\caption{Pascal VOC 2007.}
\label{fig:exp3-a}
\end{subfigure}
\begin{subfigure}[b]{0.33\textwidth}
\centering
\includegraphics[width=\linewidth]{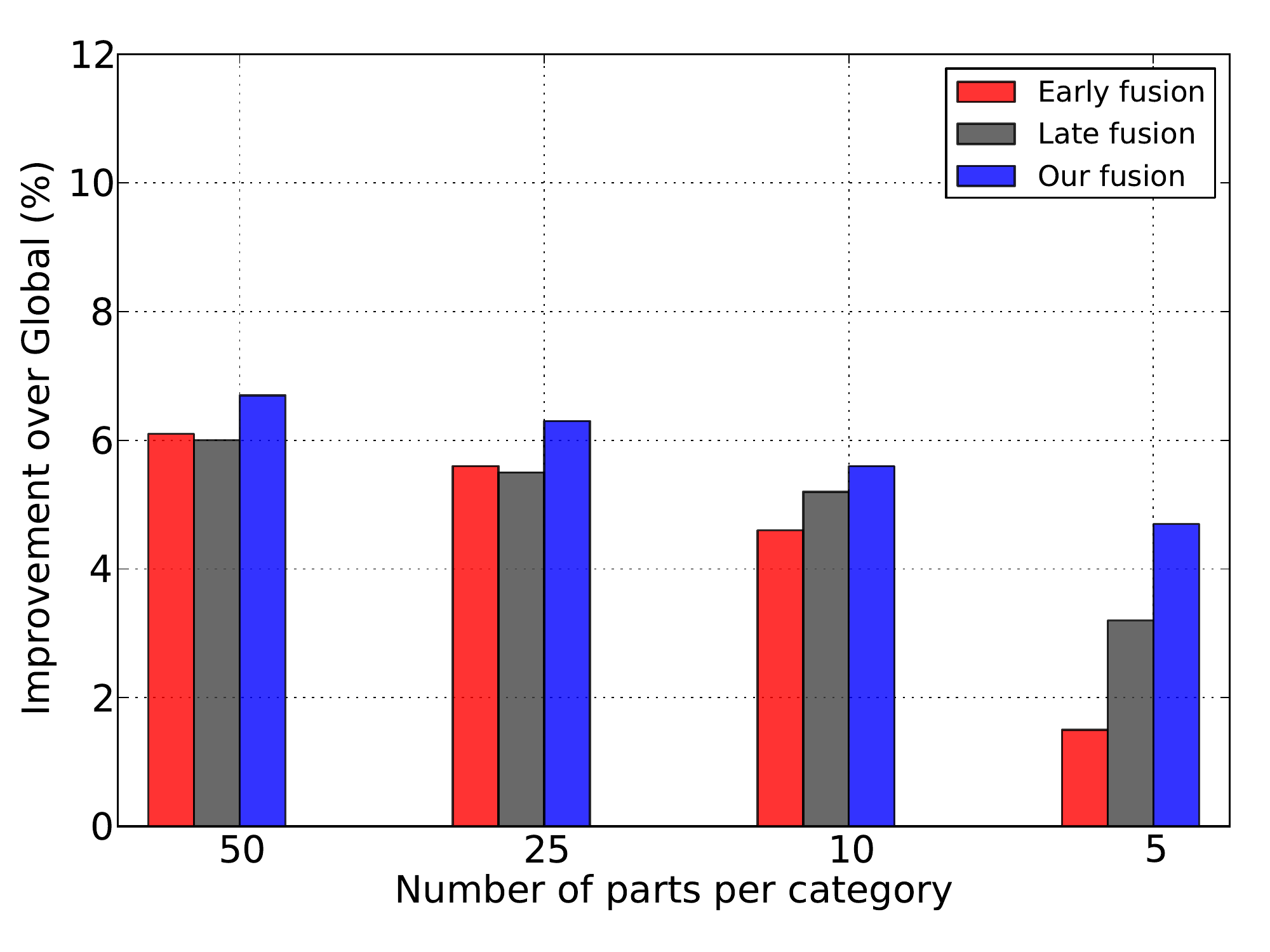}
\caption{Indoor Scenes (GoogLeNet).}
\label{fig:exp3-b}
\end{subfigure}
\begin{subfigure}[b]{0.33\textwidth}
\centering
\includegraphics[width=\linewidth]{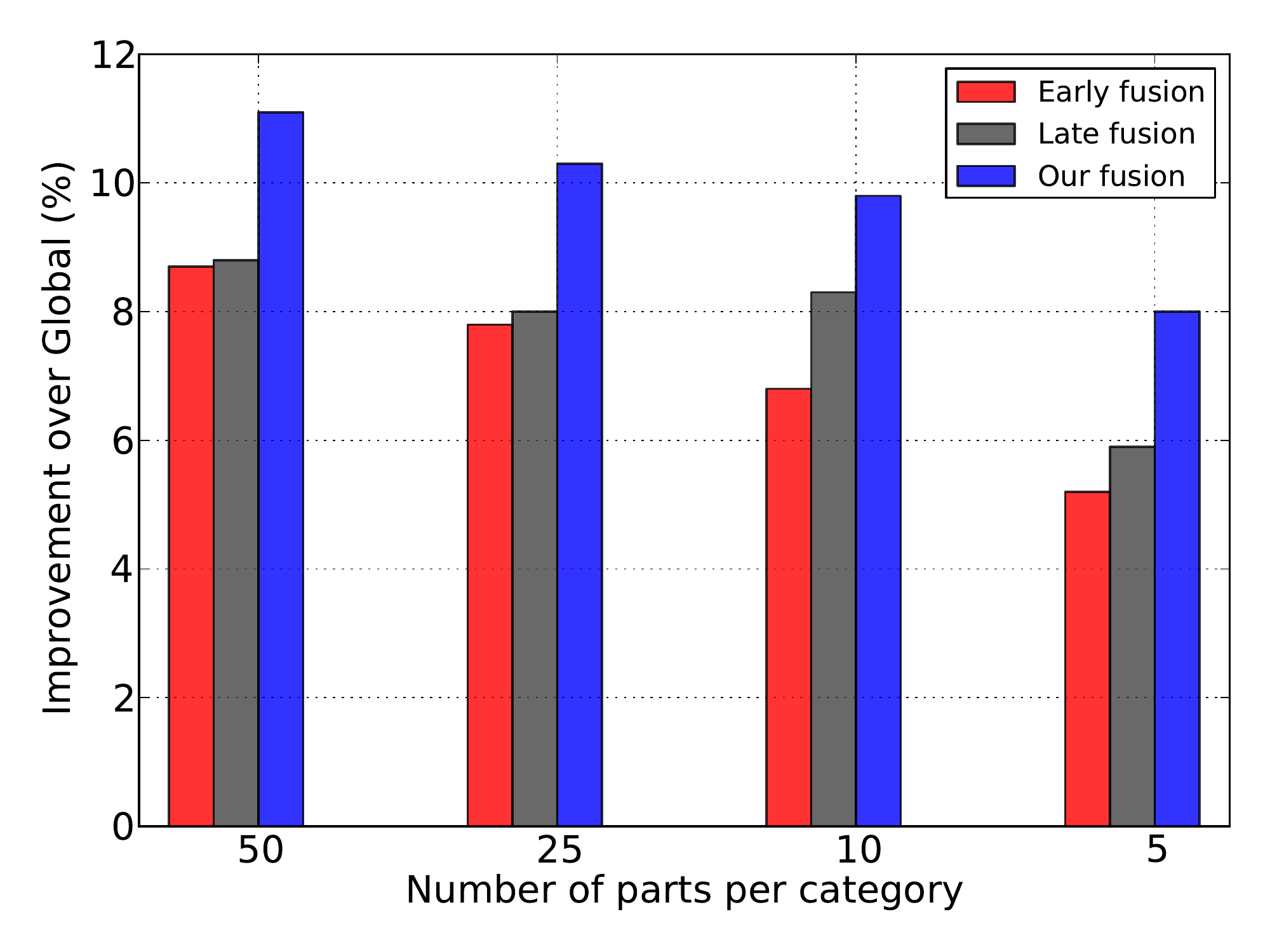}
\caption{Indoor Scenes (HOG+IFV).}
\label{fig:exp3-c}
\end{subfigure}
\caption{Combination results on Pascal VOC 2007 and MIT 67 Indoor Scenes. Ours (blue) is always favored over early (red) and late (gray) feature fusion for combining part-based and global image representations.}
\label{fig:exp3}
\end{figure*}

We also provide part responses by our method in Figure~\ref{fig:qual-exp2}. The Figure shows how our method deals with occluded objects (\ref{fig:qual-exp2-a}), small objects (
\ref{fig:qual-exp2-b}), and atypical object views (\ref{fig:qual-exp2-c}). The primary focus is on the visible portions of the objects and co-occurring other objects, but contextual information is also taken into account. Examples include car jack (\texttt{car}, \ref{fig:qual-exp2-a}), steam (\texttt{train}, \ref{fig:qual-exp2-d}), and horizontal bars (\texttt{horse}, \ref{fig:qual-exp2-g}). Our method automatically discovers that \texttt{people} parts should be shared for \texttt{dog} (\ref{fig:qual-exp2-f}), but \texttt{dog} parts should not be shared for \texttt{bike} (\ref{fig:qual-exp2-e}).
\\\\
\noindent
\textbf{Joint vs. separate selection and categorization.}
Second, we compare our method to the bag-of-parts of Juneja et al.~\cite{juneja13}, which applies a separate optimization of part selection and image categorization. To ensure a fair comparison, we replicate the setup described in~\cite{juneja13}. This means that we use the same dataset (MIT 67 Indoor Scenes), the same part proposals, and the same part features. For the HOG features used in~\cite{juneja13}, we also employ the outlined iterative part refinement. For full details, we refer to~\cite{juneja13}. We also provide a comparison using the GoogLeNet features for the parts. For the bag-of-parts, we report the results with both a linear and $\chi^{2}$ SVM, as is proposed in~\cite{juneja13}. We only exclude image flipping and spatial pyramids, as these elements are not specific to either method and can therefore cloud the comparison.

In Figure~\ref{fig:exp2-2-3-a}, we show the classification accuracies using respectively the HOG and GoogLeNet features. {\color{black}The scores the Figure are compared for four values of constraint s to evaluate the influence on the performance as a function of s for both our approach and the method of Juneja et al.~\cite{juneja13}. We enforce four global part limits: 335, 670, 1675, and 3350 parts on MIT 67 Indoor Scenes and 100, 200, 500, 1000 on Pascal VOC 2007. For the baseline bag-of-parts approach, this corresponds to 5, 10, 25, and 50 parts selected per category.}

For the HOG features, we observe that our joint part selection outperforms the separate optimization in the bag-of-parts of Juneja et al.~\cite{juneja13}. The difference even grows as we use stricter part limits. At 5 parts per category, our method outperforms bag-of-parts by 7.6\% ($\chi^{2}$) and 12.6\% (linear). For the GoogLeNet features, we first observe that all results significantly outperform the HOG features (note the different ranges on the y-axis). We again outperform the bag-of-parts, albeit with a less pronounced difference. At five parts per category, we outperform the bag-of-parts by 3.6\% ($\chi^{2}$) and 5.6\% (linear). We attribute this to the strength of the features, resulting in saturation.

The run time of our approach is similar to other part-­based methods, as they all compute part responses and apply learned classifiers. Naturally, the training time of our joint selection and classification is more involved than methods with separate selection and classification~\cite{doersch13,juneja13}, yet fast enough to avoid dimensionality reduction, as in~\cite{parizi15}. On a single Intel Xeon E5­2609 core, it takes roughly 8 hours to jointly optimize all 20 objects in Pascal VOC 2007. On a test image, it takes 2 seconds to max-­pool the selected parts and apply the trees.

We conclude that our joint part selection and categorization is preferred over a separate optimization.
\\\\
\noindent
\textbf{Evaluating maximum exploiting.}
Third, we evaluate the maximum exploiting itself within the joint AdaBoost optimization. We reuse the HOG settings of~\cite{juneja13} and draw a comparison to three baseline sampling strategies. The first baseline, \emph{Tieu-Viola}, follows~\cite{tieu04}. At each boosting iteration, a single new part for each category is selected until the specified part limit is reached. The second baseline, \emph{uniform}, is a variant of LazyBoost~\cite{escudero00}, where we randomly sample parts from the set of all parts (ignoring whether parts have been selected or not), until the limit is reached, after which we keep retraining on the selected parts. The third baseline follows the $\epsilon$-greedy strategy, where we have a static probability $\epsilon$ of exploring new parts and a static probability $1 - \epsilon$ of exploiting the current set of selected parts.

The comparison of maximum exploit sampling versus the baseline boosting sampling methods is shown in Figure~\ref{fig:exp2-2-3-b}. The maximum exploiting joint optimization is preferable across the whole range of part limits. These results indicate that a higher focus on exploitation is an important aspect in the joint optimization, since the baselines have a higher focus on exploration. It is also interesting to note that, with the exception of the \emph{Tieu-Viola} baseline, all variants improve over Juneja et al.~\cite{juneja13}, indicating the effectiveness of a joint optimization.

\subsubsection*{Experiment 3: Evaluating bootstrap fusion.}

We evaluate our bootstrap fusion for combining our category parts with the global representation. We perform this evaluation on both Pascal VOC 2007 and MIT 67 Indoor Scenes and provide a comparison to two baseline fusions~\cite{snoek05}. For the bootstrap procedure, we train a model on the global representations and then train our joint part selection and image categorization on the updated weights. The combination is then formed by concatenating the decision trees from both models. The first baseline, early fusion, combines the global representations and the representations from the parts selected by~\cite{juneja13}, after which a $\chi^{2}$ SVM is trained on the concatenated representations. The second baseline, late fusion, is similar to the bootstrap procedure, but without updating the weights.
{\color{black}In the remaining experiments we always use an average of 50 parts per category for our approach.}
\\\\
\textbf{Results.}
In Figure~\ref{fig:exp3} we show the performance of our bootstrap fusion against the two baselines. Across all settings and part limits, our fusion (blue bars) outperforms the baselines. For Pascal VOC 2007 our method outperforms the other combiners between 2-4\% across all part limits, with an mAP up to 90.7\% at 50 category parts per object. For MIT 67 Indoor Scenes, the difference in performance varies for both the GoogLeNet and HOG features. For the HOG features with the Fisher Vector as the global representation, the improvement is most significant, similar to the results of Figure~\ref{fig:exp2-2-3-a}. For the GoogLeNet features the improvement is less significant, but our bootstrap is still better than the baselines. We conclude that our bootstrap procedure is preferred when combining the part and global representations.
\begin{figure}[t]
\centering
\begin{subfigure}[b]{0.45\linewidth}
\centering
\includegraphics[width=\linewidth]{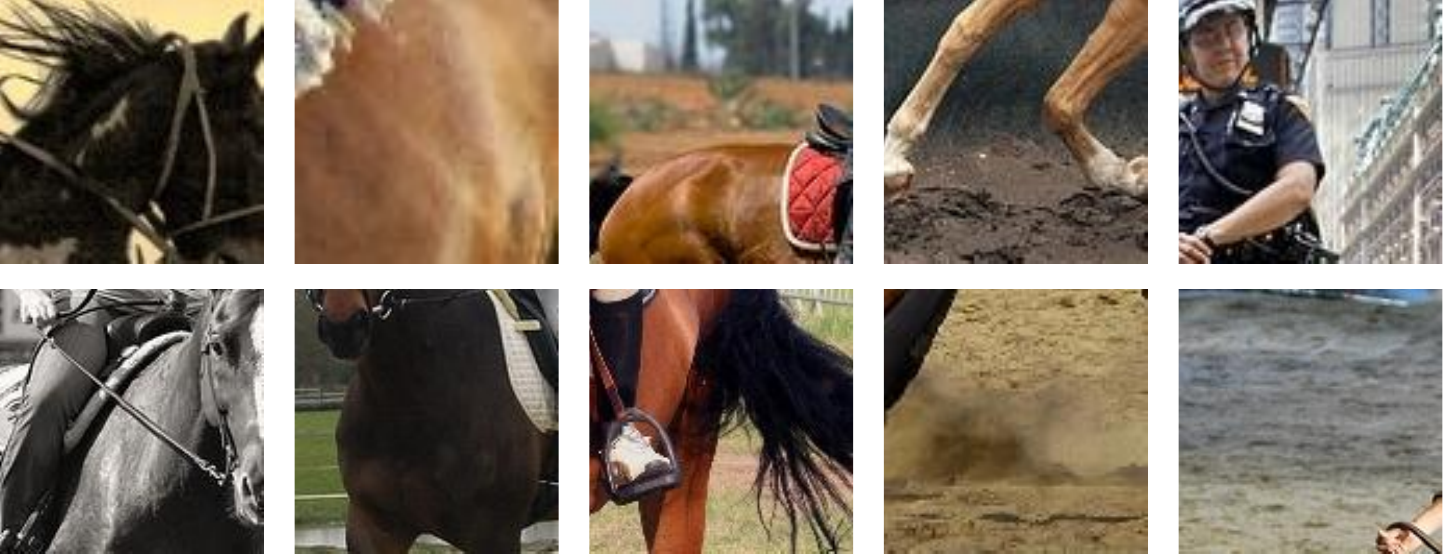}
\caption{Horse riding.}
\label{fig:willow1}
\end{subfigure}
\hspace{0.25cm}
\begin{subfigure}[b]{0.45\linewidth}
\centering
\includegraphics[width=\linewidth]{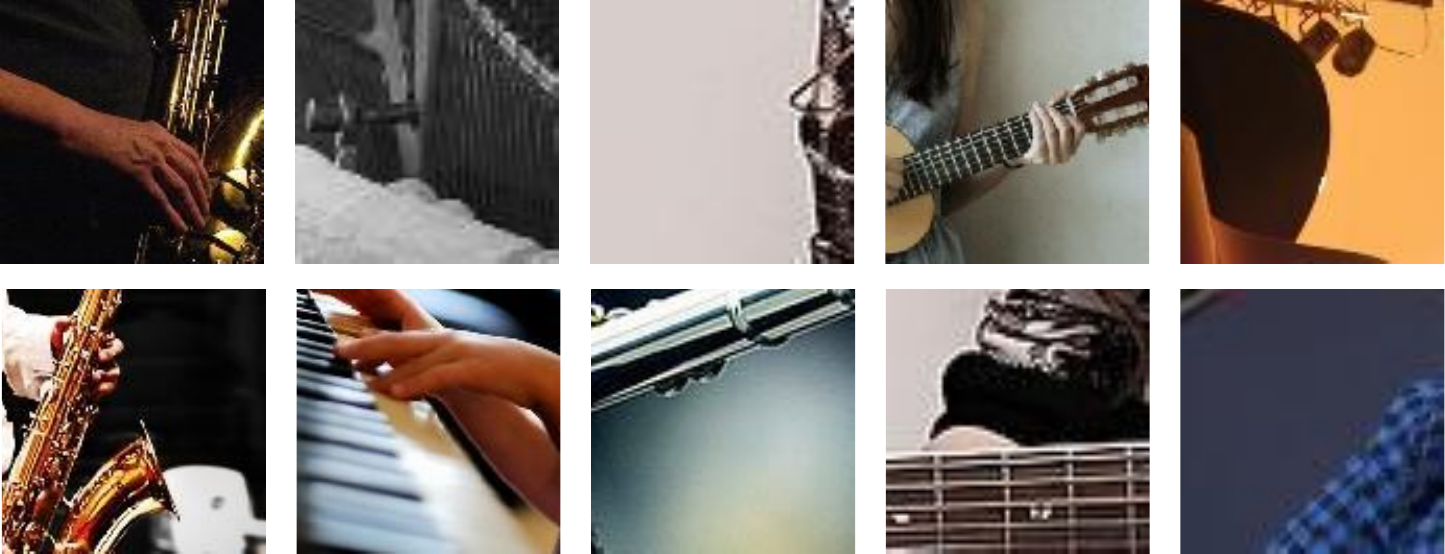}
\caption{Playing instrument.}
\label{fig:willow2}
\end{subfigure}
\caption{Selected parts from the action categories horse riding and playing music in the Willow image actions dataset. Note the presence of \emph{Own}, \emph{Other} and \emph{Context} parts.}
\label{fig:willow}
\end{figure}
\\\\
\begin{figure}[t]
\centering
\includegraphics[width=0.75\textwidth]{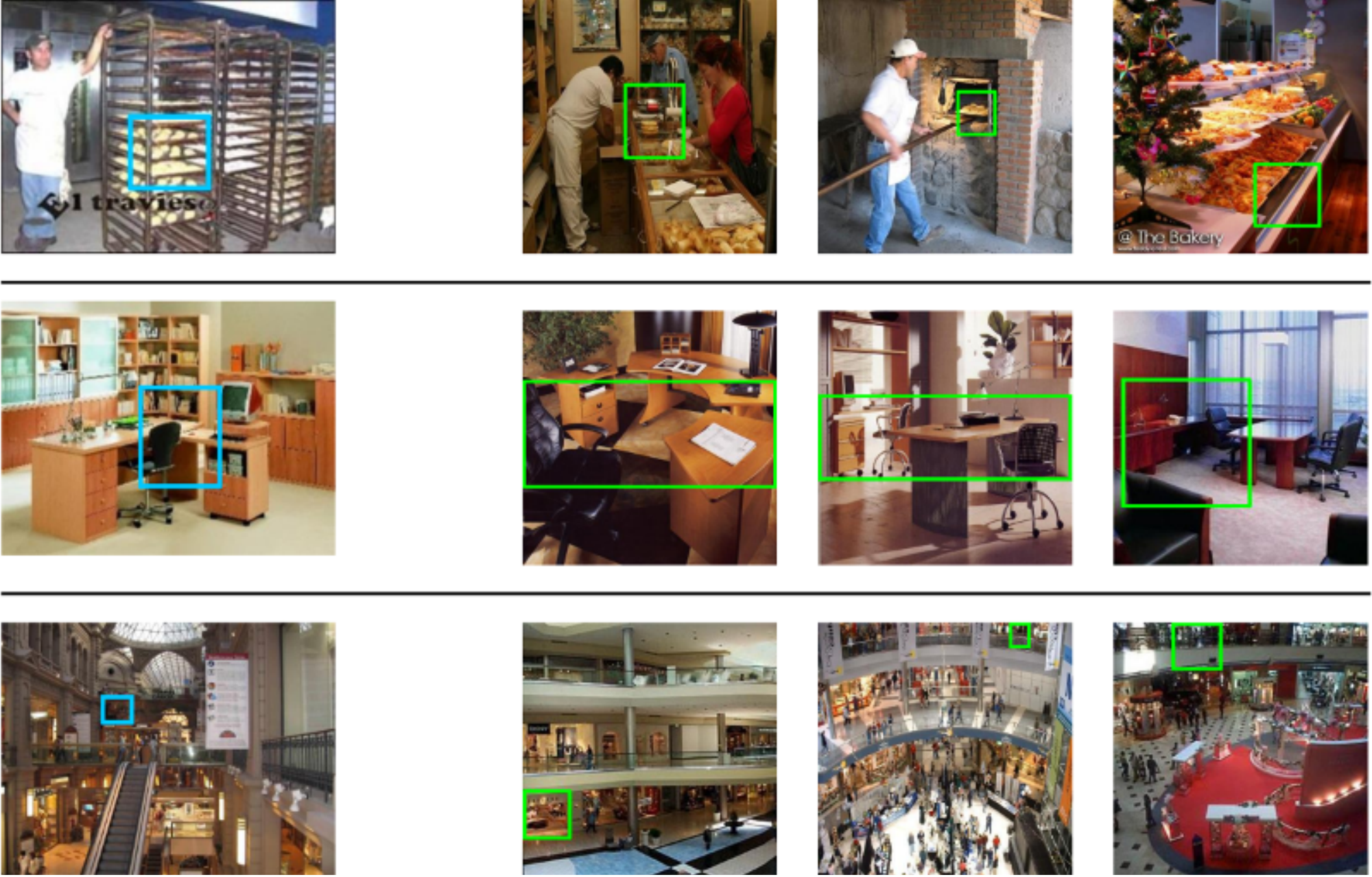}
\caption{​Three examples of selected parts (blue, left) and their top responses in three test images (green, right) on MIT Indoor Scenes. Part responses focus on similar parts of the scene (e.g. broad for bakery and shop for mall) and even on larger combinations of scene parts (e.g. desk and chair for office).}
\label{fig:qual-mit}
\end{figure}
\textbf{Comparison to the state-of-the-art.}
We also compare our results for the fusion to related work. The comparison of our results to the state-of-the-art on Pascal VOC 2007 is provided in Table~\ref{tab:exp3}. Our results provide an improvement over the references. Our final result compares favorably to the convolutional neural network applied to the whole image (90.7\% vs. 85.3\%), a notable improvement in mAP. We note that recent concurrent work on (very deep) convolutional networks similar in spirit to~\cite{szegedy14} has yielded an mAP of 89.3\%~\cite{simonyan15}, which is a 4\% improvement over the network that we employ. Both our part-based and combined results still outperform~\cite{simonyan15} and we expect a further improvement when employing their network to represent the parts. In Figure~\ref{fig:qual-mit}, we highlight 3 selected parts during training on MIT 67 Indoor Scenes, along with top responses in three test images.

The comparison for MIT 67 Indoor Scenes and Willow Actions is also provided in Table~\ref{tab:res-exp4}. For scenes, we see that our results improve upon existing combinations of global and part-based representations~\cite{doersch13,juneja13}. Similar to Pascal VOC 2007, we note that the recently introduced scene-specific convolutional network of Zhou~et al.~\cite{zhou14} has shown to be effective for global scene classification, with an accuracy of 70.8\%. Subsequent analysis on this network has indicated that convolutional layers contain object detectors~\cite{zhou15}. This analysis not only speaks in favor of the network for global classification, but also in our part-based setting, as this will positively influence the part representations.

Compared to Cimpoi et al.~\cite{cimpoi15}, we note that their results are better on MIT 67 Indoor Scenes, while we perform better on Pascal VOC 2007. This result is not surprising, as Pascal VOC 2007 is a multi-label dataset and Indoor Scenes is a multi-class dataset. As a result, the objects in Pascal VOC 2007 have a higher level of co-occurrence, which is exploited by our method.

\begin{table*}[t]
\centering
\footnotesize
\scalebox{0.725}{
\begin{tabular}{l c c c c c c c c c c c c c c c c c c c c c}
\toprule
& \multicolumn{21}{c}{\textbf{Pascal VOC 2007}}\\
\textbf{Method}
&
\begin{minipage}{0.03\textwidth} \includegraphics[width=\linewidth]{images/icons/aeroplane.png} \end{minipage}
&
\begin{minipage}{0.03\textwidth} \includegraphics[width=\linewidth]{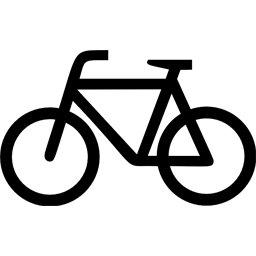} \end{minipage}
&
\begin{minipage}{0.03\textwidth} \includegraphics[width=\linewidth]{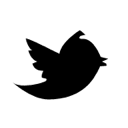} \end{minipage}
&
\begin{minipage}{0.03\textwidth} \includegraphics[width=\linewidth]{images/icons/boat.png} \end{minipage}
&
\begin{minipage}{0.03\textwidth} \includegraphics[width=\linewidth]{images/icons/bottle.png} \end{minipage}
&
\begin{minipage}{0.03\textwidth} \includegraphics[width=\linewidth]{images/icons/bus.png} \end{minipage}
&
\begin{minipage}{0.03\textwidth} \includegraphics[width=\linewidth]{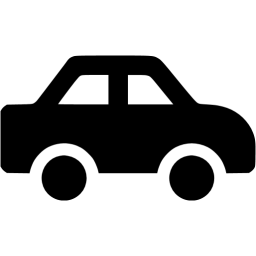} \end{minipage}
&
\begin{minipage}{0.03\textwidth} \includegraphics[width=\linewidth]{images/icons/cat.png} \end{minipage}
&
\begin{minipage}{0.03\textwidth} \includegraphics[width=\linewidth]{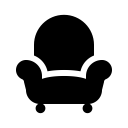} \end{minipage}
&
\begin{minipage}{0.03\textwidth} \includegraphics[width=\linewidth]{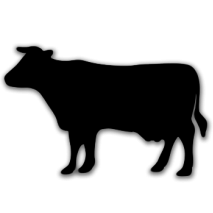} \end{minipage}
&
\begin{minipage}{0.03\textwidth} \includegraphics[width=\linewidth]{images/icons/table.png} \end{minipage}
&
\begin{minipage}{0.03\textwidth} \includegraphics[width=\linewidth]{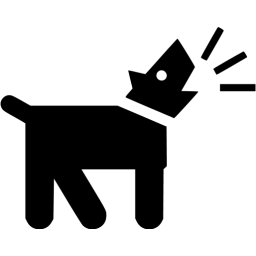} \end{minipage}
&
\begin{minipage}{0.03\textwidth} \includegraphics[width=\linewidth]{images/icons/horse.png} \end{minipage}
&
\begin{minipage}{0.03\textwidth} \includegraphics[width=\linewidth]{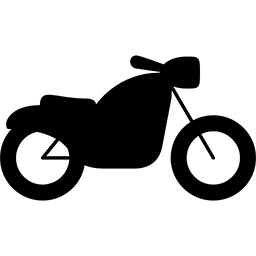} \end{minipage}
&
\begin{minipage}{0.03\textwidth} \includegraphics[width=\linewidth]{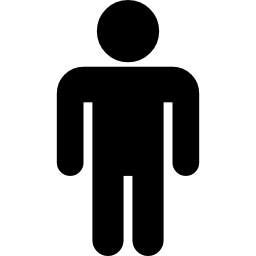} \end{minipage}
&
\begin{minipage}{0.03\textwidth} \includegraphics[width=\linewidth]{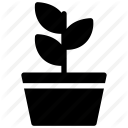} \end{minipage}
&
\begin{minipage}{0.03\textwidth} \includegraphics[width=\linewidth]{images/icons/sheep.png} \end{minipage}
&
\begin{minipage}{0.03\textwidth} \includegraphics[width=\linewidth]{images/icons/sofa.png} \end{minipage}
&
\begin{minipage}{0.03\textwidth} \includegraphics[width=\linewidth]{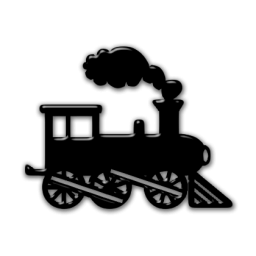} \end{minipage}
&
\begin{minipage}{0.03\textwidth} \includegraphics[width=\linewidth]{images/icons/tv.png} \end{minipage}
&
mAP
\\
\midrule
Oquab et al.~\cite{oquab14} & 88.5 & 81.5 & 87.9 & 82.0 & 47.5 & 75.5 & 90.1 & 87.2 & 61.6 & 75.7 & 67.3 & 85.5 & 83.5 & 80.0 & 95.6 & 60.8 & 76.8 & 58.0 & 90.4 & 77.9 & 77.7\\
Chatfield et al.~\cite{chatfield14} & 95.3 & 90.4 & 92.5 & 89.6 & 54.4 & 81.9 & 91.5 & 91.9 & 64.1 & 76.3 & 74.9 & 89.7 & 92.2 & 86.9 & 95.2 & 60.7 & 82.9 & 68.0 & 95.5 & 74.4 & 82.4\\
Wei et al.~\cite{wei14} & 96.0 & 92.1 & 93.7 & 93.4 & 58.7 & 84.0 & 93.4 & 92.0 & 62.8 & 89.1 & 76.3 & 91.4 & 95.0 & 87.8 & 93.1 & 69.9 & 90.3 & 68.0 & 96.8 & 80.6 & 85.2\\
Cimpoi et al.~\cite{cimpoi15} & 91.4 & 90.9 & 91.2 & 88.9 & 66.7 & 85.8 & 91.1 & 90.7 & 71.7 & 80.1 & 82.4 & 90.4 &	91.0 & 89.3 & 94.4 & 68.7 & 84.2 & 79.0 & 93.8 & 82.2 &	85.2\\
Szegedy et al.~\cite{szegedy14} & 97.7 & 93.1 & 94.4 & 93.8 & 55.1 & 87.8 & 91.6 & 95.3 & 66.1 & 78.7 & 79.6 & 93.2 & 94.4 & 90.0 & 95.4 & 64.2 & 83.6 & 74.4 & 96.4 & 80.9 & 85.3\\
Simonyan et al.~\cite{simonyan15} & & & & & & & & & & & & & & & & & & & & & 89.3\\
\midrule
Ours & 98.0 & 96.8 & 97.9 & 94.7 & 76.9 & 89.6 & 96.4 & 96.8 & 74.9 & 82.9 & 79.7 & 94.4 & 94.6 & 93.6 & 98.3 & 74.9 & 91.5 & 75.4 & 96.7 & 90.2 & \textbf{89.7}\\
Ours + ~\cite{szegedy14} & 98.7 & 97.0 & 97.9 & 94.8 & 78.3 & 91.4 & 96.4 & 97.3 & 75.0 & 85.0 & 82.4 & 95.4 & 96.1 & 94.7 & 98.5 & 75.9 & 90.9 & 82.1 & 97.3 & 89.7 & \textbf{90.7}\\
\bottomrule
\end{tabular}
}
\caption{Average Precision (\%) scores on Pascal VOC 2007, comparing our results to related work. Our part-based results improve upon existing global representations (from 85.3\% for~\cite{szegedy14} to 89.7\%), while the combination with~\cite{szegedy14} yields the best results.}
\label{tab:exp3}
\end{table*}

\begin{table}[t]
\scriptsize
\centering
\begin{tabular}{l r}
\toprule
\multicolumn{2}{c}{\textbf{MIT 67 Indoor Scenes}}\\
Method & Acc.\\
\midrule
Juneja et al.~\cite{juneja13} & 63.1\%\\
Doersch et al.~\cite{doersch13} & 66.9\%\\
Szegedy et al.~\cite{szegedy14} & 69.3\%\\
Zhou~et al.~\cite{zhou14} & 70.8\%\\
Zuo et al.~\cite{zuo14} & 76.2\%\\
Parizi et al.~\cite{parizi15} & 77.1\%\\
Cimpoi et al.~\cite{cimpoi15} & \textbf{81.0\%}\\
\midrule
Ours & 77.4\%\\
\bottomrule
\end{tabular}
\hspace{0.25cm}
\begin{tabular}{l r}
\toprule
\multicolumn{2}{c}{\textbf{Willow Actions}}\\
Method & mAP\\
\midrule
Pintea et al.~\cite{pintea14} & 51.0\%\\
Sicre et al.~\cite{sicre14} & 61.4\%\\
Delaitre et al.~\cite{delaitre10} & 62.9\%\\
Sharma et al.~\cite{sharma12} & 65.9\%\\
Sharma et al.~\cite{sharma13} & 67.6\%\\
Khan et al.~\cite{khan13} & 70.1\%\\
Szegedy et al.~\cite{szegedy14} & 74.4\%\\
\midrule
Ours & \textbf{81.7\%}\\
\bottomrule
\end{tabular}
\hspace{0.25cm}
\begin{tabular}{l r r}
\toprule
\multicolumn{3}{c}{\textbf{Calltech-UCSD Birds 200-2011}}\\
Method & \multicolumn{2}{c}{Accuracy}\\
&  w/o fine-tuning & fine-tuning\\
\midrule
Zhang et al.~\cite{zhang2014part} & 66.0\% & 73.9\%\\
Simon and Rodner~\cite{simon2015neural} & - & 81.0\%\\
Wang et al.~\cite{wang2015multiple} & - & 81.7\%\\
Krause et al.~\cite{krause2015fine} & 78.8\% & 82.0\%\\
Lin et al.~\cite{lin2015bilinear} & 80.1\% & \textbf{84.1\%}\\
Jaderberg et al.~\cite{jaderberg2015spatial} & - & \textbf{84.1\%}\\
\midrule
Ours & \textbf{82.8\%} & -\\
\bottomrule
\end{tabular}
\caption{Comparison with state-of-the-art on MIT 67 Indoor Scenes, Willow Actions, and Calltach-UCSD Birds 200-2011. Here, ft denotes fine-tuning. Using a single feature only~\cite{szegedy14}, our approach is competitive on all datasets and best for Pascal Objects and Willow Actions and Fine-grained Birds without fine-tuning.}
\label{tab:res-exp4}
\end{table}

We perform an evaluation on Willow Actions, where we see that our final result of 81.7\% outperforms current methods on the same dataset. In Figure~\ref{fig:willow} we show selected part for two action categories. The Figure paints a similar picture as the previous experiments. For both action categories, our method selects parts from the dominant objects in the scene (horses and instruments), but also from contextual information. For \texttt{horse riding}, parts such as sand and policemen are deemed discriminative, while parts from \texttt{playing music} include lumberjack prints from clothing.

{\color{black}Finally, we also provide results for the fine-grained Caltech-UCSD Birds 200-2011 dataset in Table 2. In this experiment, we exclude the use of bounding box information and only rely on the image class labels. For the results in Table 2, we use 10,000 parts (50 parts on average per category). The results show that our approach is competitive to the current state-of-the-art without the need for bounding box supervision and without any fine-grained optimizations. For example,~\cite{krause2015fine,lin2015bilinear,zhang2014part} fine-tune their deep networks to the fine-grained images during training, which delivers a 4-5\% improvement. We yield the highest scores when no fine-tuning is applied and expect a similar gain as~\cite{krause2015fine,lin2015bilinear,zhang2014part} when fine-tuning is applied.}

From this evaluation, we conclude that our method improves upon global ConvNet image representations and yields competitive results compared to the current state-of-the-art.

\section{Conclusions}
\label{sec:conclusions}
We have investigated image categorization using a representation of parts. We start from the intuition that parts are naturally shared between categories. We analyze the type of region where a part comes from, namely: parts from its \emph{own} category, parts from \emph{other} categories, and \emph{context} parts. Experimental validation confirms that all the types of parts contain valuable and complementary information. To share parts between categories we extend Adaboost, profiting from joint part selection and image classification. Experimental evaluation on object, scene, action, and fine-grained categories shows that our method is capable of incorporating all three types of parts. Furthermore, our method provides a further boost in performance over a convolutional representation of the whole image for categorization.
{\color{black}The strong performance of our and other state-of-the-art methods on Pascal VOC 2007, MIT 67 Indoor Scenes, and Calltech-UCSD Birds 200-2011 indicates the saturation on these datasets and the need for the community to explore bigger datasets for part-based image categorization in the near future. Our method opens up new possibilities for part-based methods, such as image captioning by sharing parts over similar words within the captions.}

\section*{Acknowledgements}
This research is supported by the STW STORY project and the Dutch national program COMMIT.

\section*{References}
\bibliographystyle{abbrv}
\bibliography{parts-cviu}

\end{document}